\documentclass[letterpaper]{article}
\usepackage[preprint]{aaai2027}
\usepackage[hyphens]{url}
\usepackage{natbib}
\usepackage{multibib}
\newcites{app}{Appendix References}
\usepackage{caption}
\usepackage{subcaption}
\usepackage[utf8]{inputenc}
\usepackage[T1]{fontenc}
\usepackage{microtype}
\usepackage{booktabs}
\usepackage{amsfonts}
\usepackage{amsmath}
\usepackage{amssymb}
\usepackage{nicefrac}
\usepackage{xcolor}
\usepackage{colortbl}
\usepackage{graphicx}
\usepackage{tikz}
\usepackage{multirow}
\usepackage{makecell}
\usepackage{listings}
\usepackage{algorithm}
\usepackage{algpseudocode}
\usepackage[most]{tcolorbox}
\usepackage{pifont}
\tcbuselibrary{listings,breakable}
\usetikzlibrary{arrows.meta,positioning,fit,calc}

\newcounter{prompt}[section]
\renewcommand{\theprompt}{\thesection.\arabic{prompt}}
\newtcblisting{promptbox}[2][]{
  listing only,
  enhanced, breakable,
  phantom={\refstepcounter{prompt}},
  colback=blue!2!white,
  colframe=blue!25!black!40,
  colbacktitle=blue!8!white,
  coltitle=black,
  fonttitle=\footnotesize\bfseries,
  title={Prompt~\theprompt: #2},
  arc=1.2mm, boxrule=0.5pt, titlerule=0.3pt,
  top=3pt, bottom=3pt, left=4pt, right=4pt,
  before skip=8pt, after skip=8pt,
  listing options={
    basicstyle=\ttfamily\scriptsize,
    breaklines=true,
    breakatwhitespace=false,
    breakindent=0pt,
    breakautoindent=false,
    columns=fullflexible,
    keepspaces=true,
    upquote=true,
    frame=none,
    xleftmargin=0pt,
    xrightmargin=0pt,
    aboveskip=0pt,
    belowskip=0pt,
    literate={"}{\textquotedbl}1
  },
  #1
}

\newcommand{\partialmark}{\tikz[baseline=-0.65ex]{\draw (0,0) circle (0.75ex); \fill (0,0.75ex) arc (90:270:0.75ex) -- cycle;}}
\title{Setoka: A Benchmark for Hierarchical User Understanding in Personalized Agents over Heterogeneous Data}
\author{
    Lingyang Zeng\textsuperscript{\rm 1},
    Guangze Chen\textsuperscript{\rm 1},
    Kaichen Yu\textsuperscript{\rm 1},
    Zhicheng Pan\textsuperscript{\rm 2},
    Siyang Weng\textsuperscript{\rm 1},
    Zirui Hu\textsuperscript{\rm 1},
    \mbox{Xiangyun Du}\textsuperscript{\rm 1},
    Hailin He\textsuperscript{\rm 1},
    Rong Zhang\textsuperscript{\rm 1},
    Chengcheng Yang\textsuperscript{\rm 1},
    Kai Huang\textsuperscript{\rm 1},
    Xuan Zhou\textsuperscript{\rm 1}
}
\affiliations{
    \textsuperscript{\rm 1}East China Normal University\\
    \textsuperscript{\rm 2}The Chinese University of Hong Kong\\
    \texttt{\{lyzeng, gzchen, kaichenyu, syweng, zrhu, dxy, hlhe\}@stu.ecnu.edu.cn},
    \texttt{zcpan@se.cuhk.edu.hk},\\
    \texttt{\{rzhang, ccyang, khuang, xzhou\}@dase.ecnu.edu.cn}
}

\begin{document}

\maketitle

\begin{abstract}
Personalized agents are increasingly applied to assist users across a wide range of tasks.
Effective personalized assistance requires not only retrieving explicit facts from past interactions stored in agent memory, but also inferring abstract personal characteristics.
However, existing memory benchmarks primarily evaluate whether an agent can retrieve information explicitly stated in conversational histories, failing to provide an effective assessment of deeper user understanding.
In this work, we propose \textsc{Setoka}, a benchmark for evaluating memory-augmented personalized agents with hierarchical user understanding from heterogeneous data.
Grounded in theories from cognitive and personality psychology, \textsc{Setoka} defines four levels of user understanding, i.e., \textit{semantic memory}, \textit{episodic memory}, \textit{behavior pattern}, and \textit{personality trait}.
Moreover, to enable realistic yet privacy-preserving evaluation, we design a psychometrics-based pipeline that synthesizes diverse, coherent heterogeneous user data and queries at scale.
Finally, we leverage \textsc{Setoka} to evaluate 3 language models combined with 5 memory systems for 10 synthetic users.
Our comprehensive evaluation reveals that while existing systems perform well on semantic memory retrieval, their performance declines on episodic memory.
Moreover, when dealing with behavior pattern and personality trait understanding tasks that require integrating heterogeneous and fragmented information dispersed over time, performance declines even further.
These findings demonstrate that user understanding cannot be handled by simple fact retrieval, motivating the design of memory mechanisms for cross-source integration and abstraction over long-term user behavior.

\end{abstract}

\section{Introduction}

Large language model (LLM)-based agents are increasingly deployed as personalized assistants~\cite{li2024personalllmagentsinsights} that understand user instructions and autonomously execute them.
The great advantage of such agents stems from their ability to handle tasks with varying degrees of user understanding.
Generally, some tasks require only retrieving an explicit fact (e.g., a scheduled meeting time), while more complicated tasks demand a nuanced comprehension that captures the user's behavioral patterns and personality traits.
However, such {abstract} characteristics are rarely stated explicitly.
Instead, they are distributed across heterogeneous data sources, including unstructured text (e.g., messages and notes), structured records (e.g., contacts and calendars), and graphs (e.g., social connections)~\cite{hu2026memoryageaiagents}.
For instance, the personality trait ``\textit{extraversion}'' cannot be inferred from any single source alone.
It requires jointly considering frequent group events in the calendar, active engagement in group chats, and numerous ties in the social graph.
This distributed nature makes straightforward retrieval-based approaches ineffective, as no individual data source provides a comprehensive understanding of the user.

Recent years have witnessed rapid advances in memory systems for agents~\cite{chhikara2025mem0}.
Benchmarks for memory-augmented agents serve as a critical touchstone for measuring these advances, revealing potential limitations and guiding the development of more capable memory systems.
However, existing benchmarks ~\cite{maharana2024locomo,wu2025longmemeval,du2024perltqa,tan2025membench,jiang2025personamem,jiang2025personamemv2,wu2026knowmebench} mainly assess the retrieval accuracy of facts explicitly stated in conversational histories.
Consequently, they provide insufficient insight into whether memory systems could integrate implicit evidence across heterogeneous data sources to infer abstract personal characteristics.

However, building such a benchmark is non-trivial due to two primary reasons.
First, collecting heterogeneous data from real-world users raises privacy concerns, and acquiring data at the scale required for systematic evaluation is also prohibitively costly.
Second, evaluating deep user understanding requires grounding answers in a coherent profile, where high-level traits are substantiated by underlying records.
For clarity, the specific challenges are summarized as follows.

\vspace{1mm}\textbf{Hierarchical User Understanding (C1)}.
User understanding requires inherently different reasoning operations, ranging from retrieving an explicit fact to inferring a personality trait.
Constructing such a hierarchy is non-trivial, as it requires both distinct reasoning scopes and a seamless transition from specific records to abstract characteristics.

\vspace{1mm}\textbf{Realistic Persona Sampling (C2)}.
A realistic persona should resemble a coherent individual rather than a random bundle of personality traits.
In real-world populations, personality traits do not occur in isolation.
That is, some trait combinations are common, whereas others are rarely observed.
Directly prompting an LLM to construct personas or sampling each personality trait individually might produce trait combinations that are unlikely to occur in the real world.

\vspace{1mm}\textbf{Diverse Trait-Behavior Mapping (C3)}.
The same personality trait can {lead to} diverse behavior patterns across different event categories.
However, existing data synthesis methods {that rely on} persona-conditioned LLM generation might map a personality trait profile to a narrow set of stereotypical behaviors~\cite{cheng2023marked,liu2024evaluating}, overlooking the inter-individual diversity documented in personality psychology.

\vspace{1mm} \textbf{Scalable and Consistent Generation (C4)}.
Mapping behavior patterns into long-term, heterogeneous user data introduces two coupled problems.
First, generation based solely on an LLM is difficult to scale to long histories, as earlier behavioral constraints might be forgotten or violated as the context grows.
Second, independently generating heterogeneous records might introduce contradictions between data items that describe the same underlying event.

\vspace{1mm}
To address these challenges, we propose \textsc{Setoka}, a dedicated benchmark {that enables the comprehensive evaluation of} hierarchical u\textbf{S}er und\textbf{e}rs\textbf{t}anding for pers\textbf{o}nalized tas\textbf{k}-oriented \textbf{a}gents.
Specifically, to address \textbf{C1}, we propose a \emph{four-level framework of user understanding}, grounded in cognitive and personality psychology, that comprises \textit{semantic memory} (SM), \textit{episodic memory} (EM), \textit{behavior patterns} (BP), and \textit{personality traits} (PT).
{These four levels {span} a continuum from concrete \textit{personal memories} to abstract \textit{personal characteristics}}.
For \textbf{C2}, we propose a \emph{correlation-aware personality trait sampling} strategy.
That is, the personality traits are jointly sampled from a multivariate Gaussian distribution whose covariance matrix is derived from meta-analytic psychometric correlations.
By preserving empirically observed correlations among personality dimensions, this approach produces more coherent and psychologically grounded personality trait profiles.
For \textbf{C3}, we propose \emph{psychological-scale-guided behavior pattern generation}.
The sampled personality traits are mapped to behavior patterns through validated psychological scales~\cite{soto2017bfi2}, which provide fine-grained behavior patterns grounded in established psychological constructs rather than LLM-only generation.
For \textbf{C4}, we propose an \emph{event-grounded generation tree} that organizes behavioral patterns and contexts into concrete events.
Under each parent node, child events are generated sequentially, each conditioned on its preceding siblings.
Then, their details are expanded independently in parallel without retaining the full event history.
All heterogeneous records are derived from their underlying events to maintain cross-record consistency.

In summary, we make the following contributions:

\begin{itemize}
\item We identify a fundamental limitation of existing benchmarks: they do not adequately evaluate the multiple levels of user understanding required by personalized agents.

\item {To the best of our knowledge, we introduce the first psychology-grounded framework that evaluates hierarchical user understanding, spanning concrete personal memories and abstract personal characteristics.
      By explicitly defining the evidence scope and reasoning operation required at each level, the framework enables different levels of user understanding to be evaluated independently.
      }

\item {We develop a psychometrics-based generation pipeline that samples coherent user profiles and generates long-term heterogeneous records from them at scale.
      This design provides reference answers not only for concrete personal memories, but also for abstract personal characteristics, enabling level-specific queries with traceable evidence.
      }

\item
      {We instantiate \textsc{Setoka} with 10 synthetic user profiles and 23 schemas over 3 data models.
      Then, we conduct a comprehensive evaluation of 3 language models paired with 5 popular memory systems.
      The results demonstrate that simple fact retrieval is insufficient, motivating that memory mechanisms should integrate cross-source evidence and abstract long-term user behavior.
      }

\end{itemize}

\section{Related Work}

\begin{table*}[t]
\centering
\small
\setlength{\tabcolsep}{4pt}
\begin{tabular*}{\textwidth}{@{\extracolsep{\fill}} l l c c c c r r}
\toprule
\textbf{Benchmark} & \textbf{Memory Origin} & \makecell{\textbf{Semantic}\\\textbf{Memory}} & \makecell{\textbf{Episodic}\\\textbf{Memory}} & \makecell{\textbf{Behavior}\\\textbf{Pattern}} & \makecell{\textbf{Personality}\\\textbf{Trait}} & \textbf{\#DM} & \textbf{\#Sch.} \\
\midrule
LoCoMo~\cite{maharana2024locomo}        & Dialogue          & \checkmark & \checkmark & \texttimes & \texttimes & 1 & 1 \\
LongMemEval~\cite{wu2025longmemeval}    & Dialogue          & \checkmark & \checkmark & \texttimes & \texttimes & 1 & 1 \\
PerLTQA~\cite{du2024perltqa}            & Profiles, Events, Dialogues & \checkmark & \checkmark & \texttimes & \texttimes & 1 & 4 \\
MemBench~\cite{tan2025membench}         & Dialogue          & \checkmark & \texttimes & \partialmark{} & \texttimes & 1 & 1 \\
PersonaMem~\cite{jiang2025personamem}   & Dialogue          & \checkmark & \texttimes & \texttimes & \texttimes & 1 & 1 \\
PersonaMem-v2~\cite{jiang2025personamemv2} & Dialogue       & \checkmark & \texttimes & \partialmark{} & \texttimes & 1 & 1 \\
KnowMe-Bench~\cite{wu2026knowmebench}   & Autobiographical Narratives & \checkmark & \checkmark & \texttimes & \partialmark{} & 1 & 1 \\
\midrule
\textsc{Setoka} & \textbf{Heterogeneous Data} & \checkmark & \checkmark & \checkmark & \checkmark & \textbf{3} & \textbf{23} \\
\bottomrule
\end{tabular*}
\caption{Comparison of \textsc{Setoka} with representative personalized memory benchmarks.
\#DM denotes the number of data models exposed to the agent, and \#Sch. denotes the number of distinct data schemas; counts greater than one indicate heterogeneous data.
Explicitly stated behavior patterns or personality traits in natural language such as "I usually hike on weekends" are categorized as SM because they can be directly retrieved without being inferred from underlying user data.
\partialmark{} denotes partial coverage.
}
\label{tab:benchmark_comparison}
\vspace{5pt}
\end{table*}

\subsection{Memory Systems}
As personalized assistants, agents should recall user-specific context from past interaction histories~\cite{li2024personalllmagentsinsights,hu2026memoryageaiagents}.
Recent memory-augmented agents employ diverse mechanisms to retain and use information beyond the context window.
MemGPT~\cite{packer2023memgpt} manages limited context by moving information between working memory and external storage.
Mem0~\cite{chhikara2025mem0} compresses conversations into a compact, editable store of user facts, revising or removing older facts as new information arrives.
Cognee and HippoRAG 2~\cite{markovic2025cognee,gutierrez2025hipporag2} organize memories in graph structures to support relational and multi-hop retrieval.
MemMachine~\cite{wang2026memmachine} retrieves a matched memory together with nearby memories in time, providing more complete context.
{However, memory-augmented personalized agents are still far from understanding users in depth, and therefore comprehensive evaluation is essential for further progress.
\textsc{Setoka} fills this critical gap by providing a unified benchmark for evaluating hierarchical user understanding over heterogeneous data.}

\subsection{Memory Benchmarks}
Existing memory benchmarks primarily evaluate whether models can retain information from conversational histories or textual user-profile records.
Specifically, LongMemEval and LoCoMo formulate memory as the retention of and reasoning over facts grounded in long conversational histories~\cite{wu2025longmemeval,maharana2024locomo}.
PersonaMem assesses whether models can infer evolving user profiles from multi-session histories and select responses aligned with users' current states~\cite{jiang2025personamem}.
PerLTQA focuses on question answering over semantic and episodic personal memories, including profiles, social relationships, events, and dialogues, represented within a JSON data model~\cite{du2024perltqa}.

More recent benchmarks move beyond explicitly stated facts.
Specifically, {MemBench introduces reflective memory, which requires models to synthesize interaction histories to infer users’ preferences and emotions~\cite{tan2025membench}.
PersonaMem-v2 requires inferring implicitly revealed preferences from conversational nuance~\cite{jiang2025personamemv2}.
KnowMe-Bench defines levels of person understanding ranging from factual recall to principle-level reasoning over long autobiographical narratives~\cite{wu2026knowmebench}.
Collectively, these benchmarks primarily evaluate limited user understanding ability over a homogeneous interaction stream.
}
Table~\ref{tab:benchmark_comparison} summarizes how \textsc{Setoka} differs from current memory benchmarks: it is the only benchmark that evaluates an agent's user understanding ability across heterogeneous data models rather than a single dialogue or profile source, and the only evaluation suite that covers all four levels of understanding.

\section{User Understanding Modeling}
\label{sec:benchmark}

\begin{figure*}[t]
\centering
\includegraphics[width=\linewidth]{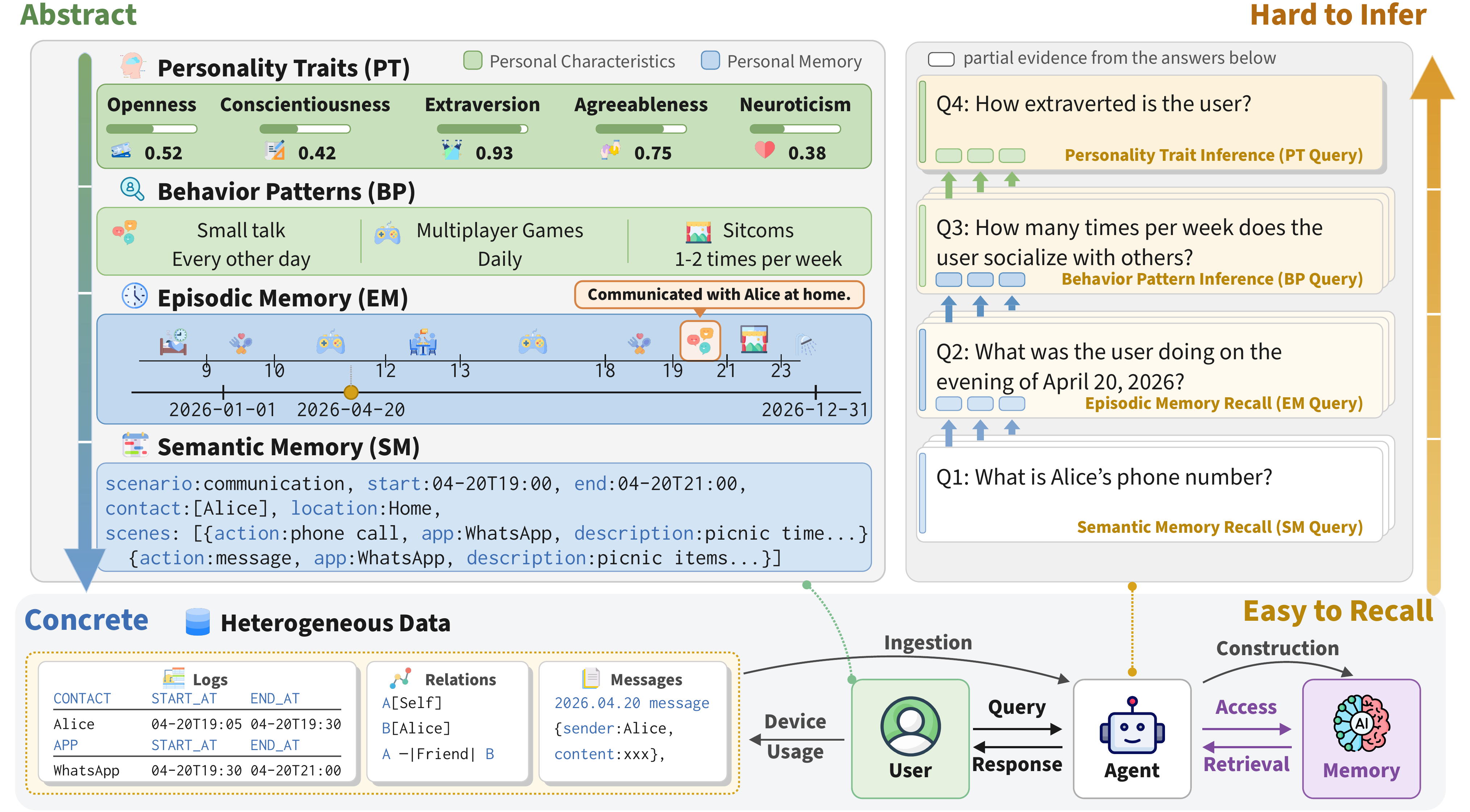}
\caption{\textbf{Overview of \textsc{Setoka}.}
\textbf{Left:} \textsc{Setoka} generates user profiles spanning from abstract personality traits to concrete semantic memory, supporting four levels of user understanding.
These profiles are then used to generate heterogeneous user data and queries at each level.
\textbf{Right:}
Representative questions at each level.
As the level of user understanding increases, answering the corresponding questions requires integrating evidence from a broader range of user data.
\textbf{Bottom:} {The generated heterogeneous records are serialized and consumed by the memory system under evaluation to construct its memory.
To answer user queries, the agent retrieves relevant information from the memory and uses it to generate responses.
}}
\label{fig:setoka-arch}
\end{figure*}

\begin{figure*}[t]
\centering
\includegraphics[width=\linewidth]{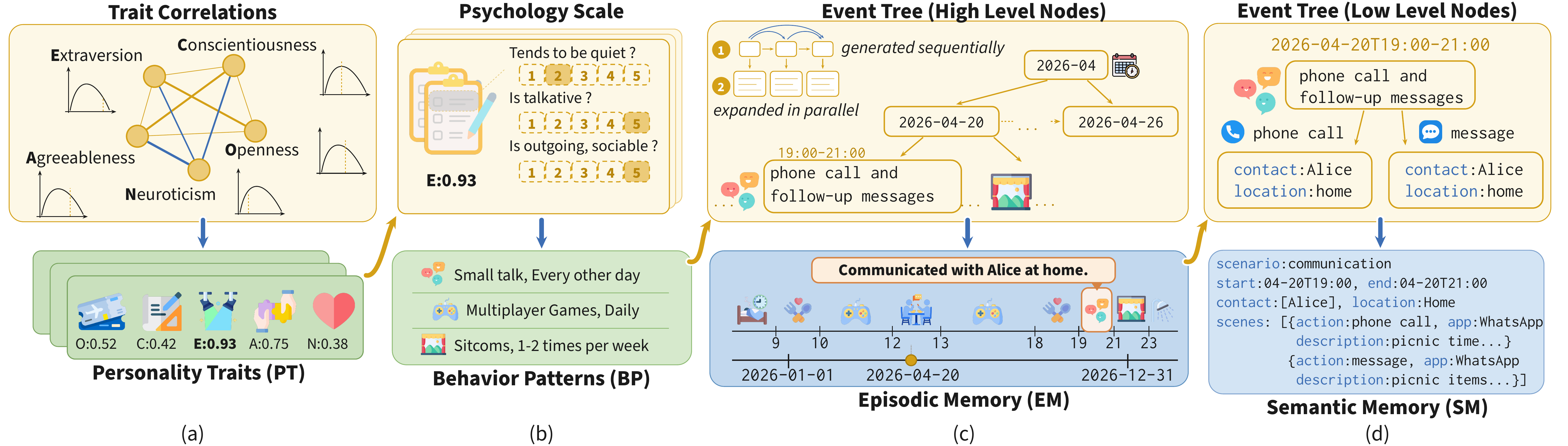}
\vspace{0.1cm}
\caption{\textsc{Setoka}'s psychometrics-based data generation pipeline.
(a) \emph{Correlation-aware personality trait sampling} draws Big-Five persona vectors jointly from a Gaussian model parameterized by meta-analytic trait correlations.
(b) \emph{Psychological-scale-based behavior pattern generation} converts each persona vector into item-level responses on a validated scale (BFI-2), yielding behavior patterns with associated frequencies.
(c) An \emph{event-grounded generation tree} expands the behavior patterns coarse-to-fine into a temporally coherent timeline of dated events.
(d) Event detail generation derives linked heterogeneous records from each event, treating the event as their single source of truth.
Only these final heterogeneous records are used by the memory system under evaluation for memory construction; the intermediate outputs serve as ground truth for query generation and evaluation.
}
\label{fig:setoka-pipeline}
\end{figure*}

\vspace{5pt}
\subsection{Problem Formulation}
\label{sec:problem-formulation}
We define \textbf{User Understanding Question Answering (UUQA)} as reasoning over a user's heterogeneous data to answer queries that explore different levels of user understanding.
\textsc{Setoka} aims to generate a realistic UUQA benchmark instance, which consists of a user's corpus $\mathcal{D}$ and a testing set $\mathcal{T}$.
Each testing instance is a tuple $(q,a,\mathcal{E})\in\mathcal{T}$, where $q$ is a testing query, $a$ is its reference answer, and $\mathcal{E}\subseteq\mathcal{D}$ is a minimal subset of records sufficient to derive $a$.
Here, the corpus is modeled as a union over a collection of heterogeneous records from multiple data resources $\mathcal{S}$: \begin{equation}
\mathcal{D} \;=\; \bigcup_{s \in \mathcal{S}} \mathcal{D}^{(s)},
\label{eq:corpus}
\end{equation} Note that, in real-world applications, $\mathcal{D}$ is typically generated through user interactions.

Following recent agentic memory formulations~\cite{hu2026memoryageaiagents}, we evaluate a memory-augmented agent with a three-stage pipeline.
Specifically, a preprocessing operator $\mathcal{P}$ first performs \emph{memory construction}, converting the raw corpus $\mathcal{D}$ into a normalized memory store: \begin{equation}
\mathcal{M} = \mathcal{P}(\mathcal{D}).
\end{equation} Then, for each query $q$ in a testing instance $(q,a,\mathcal{E})$, a retriever $\mathcal{R}$ retrieves from $\mathcal{M}$ a relevant evidence set: \begin{equation}
\hat{\mathcal{E}} = \mathcal{R}(q, \mathcal{M}).
\end{equation} Finally, an answer generator $\mathcal{G}$ performs the \emph{answer generation}, producing a response based on the query and the retrieved evidence: \begin{equation}
\hat{a} = \mathcal{G}(q, \hat{\mathcal{E}}).
\end{equation} Then, the generated response $\hat{a}$ is evaluated against its corresponding reference answer $a$ using a level-specific metric: \begin{equation}
v = \mathcal{V}_{\ell}(\hat{a}, a),
\end{equation} where $\ell$ denotes the user understanding level of the query.

\subsection{User Understanding Hierarchy}
\label{sec:levels}

User understanding spans multiple levels of personal information abstraction.
As illustrated in Fig.~\ref{fig:setoka-arch}, the four questions all concern personal information about the same user.
However, answering them requires quite different kinds of operations, ranging from direct fact retrieval to personality-trait inference.
As a result, a single aggregate score would collapse these distinctions and obscure a system's strengths and weaknesses at individual levels.
This is because a similar aggregate score might arise from significantly different performance outcomes across levels.
Therefore, we propose to use the level-specific queries and metrics to evaluate each level separately.

Inspired by cognitive and personality psychology~\cite{tulving1972episodic,mcadams1995what}, we organize user understanding into two complementary aspects, which are \textbf{personal memories} and \textbf{personal characteristics}.
The former comprises \emph{semantic memory (SM)} and \emph{episodic memory (EM)} that represent concrete personal information and experiences, while the latter comprises \emph{behavior patterns (BP)} and \emph{personality traits (PT)} that represent long-term, stable  personal  characteristics.

\vspace{5pt} \noindent \textbf{Semantic Memory (SM)} is the most concrete level of user understanding.
It consists of explicit facts about the user or the user's world, e.g., a contact's phone number.
Thus, an SM query can typically be answered by retrieving a single record.
Therefore, such evidence is explicit, localized, and directly corresponding to the requested information.

\vspace{5pt} \noindent \textbf{Episodic Memory (EM)} refers to  specific events situated {in a specific time and place}.
It captures what happened, when and where it occurred, and who was involved.
Therefore, the evidence of an EM query is distributed across related records and should be integrated.
For example, answering the query ``What was the user doing on the evening of April 20?'' requires combining calendar entries, messages, and location records that describe different aspects of the same event.

\vspace{5pt} \noindent \textbf{Behavior Pattern (BP)} represents regularities in how the user behaves across repeated occurrences of a specific event category, e.g., socializing with friends.
The evidence for a BP query is distributed over a long time span, and should be summarized to determine how the user usually behaves in a given event category.
For example, answering the query ``How many times per week does the user socialize with others?'' requires aggregating social activities across multiple weeks.

\vspace{5pt}\noindent \textbf{Personality Trait (PT)} characterizes how the user tends to think, feel, and behave across different event categories, e.g., extraversion or conscientiousness.
A PT query requires integrating evidence across multiple event categories.
For example, answering the query ``How extraverted is the user?'' might require combining evidence from social gatherings and leisure activities, rather than relying on a single type of event.

These four levels differ in how their supporting evidence is distributed and transformed.
For clarity, we formalize such transformations with four operators: \underline{selection} ($\sigma$), \underline{linking} ($\lambda$), \underline{aggregation} ($\gamma$), and \underline{generalization} ($\rho$).
Let $\mathcal{E}_{\ell}$ denote the evidence required to answer a query at level $\ell$, with the query omitted from the notation for simplicity.
Moreover, each higher level operates on a collection of outputs from lower levels, and thus its evidence set is the union of the evidence sets supporting those lower-level outputs:

\vspace{-12pt} {\small \setlength{\jot}{1pt} \begin{equation}
\begin{alignedat}{2}
a_{\mathrm{SM}}
&= \sigma(\mathcal{E}_{\mathrm{SM}}),
&&\quad |\mathcal{E}_{\mathrm{SM}}| = 1,\\
a_{\mathrm{EM}}
&= \lambda\big(\{a_{\mathrm{SM}}^{(j)}\}_{j}\big)
 = (\lambda{\circ}\sigma)(\mathcal{E}_{\mathrm{EM}}),
&&\quad
\mathcal{E}_{\mathrm{EM}}
 = \bigcup_{j}\mathcal{E}_{\mathrm{SM}}^{(j)},\\
a_{\mathrm{BP}}
&= \gamma\big(\{a_{\mathrm{EM}}^{(i)}\}_{i}\big)
 = (\gamma{\circ}\lambda{\circ}\sigma)(\mathcal{E}_{\mathrm{BP}}),
&&\quad
\mathcal{E}_{\mathrm{BP}}
 = \bigcup_{i}\mathcal{E}_{\mathrm{EM}}^{(i)},\\
a_{\mathrm{PT}}
&= \rho\big(\{a_{\mathrm{BP}}^{(c)}\}_{c}\big)
 = (\rho{\circ}\gamma{\circ}\lambda{\circ}\sigma)(\mathcal{E}_{\mathrm{PT}}),
&&\quad
\mathcal{E}_{\mathrm{PT}}
 = \bigcup_{c}\mathcal{E}_{\mathrm{BP}}^{(c)}.
\end{alignedat}
\label{eq:hierarchy}
\end{equation}}

\vspace{-3pt}

At the SM level, $\sigma(\mathcal{E}_{\mathrm{SM}})$ retrieves a fact explicitly stated in a single record.
{At the EM level, $\lambda\big(\{a_{\mathrm{SM}}^{(j)}\}_{j}\big)$ links multiple SM-level answers extracted from records describing the same event.
At the BP level, $\gamma\big(\{a_{\mathrm{EM}}^{(i)}\}_{i}\big)$ aggregates multiple EM-level answers for events within the same event category to identify a behavior pattern.
At the PT level, $\rho\big(\{a_{\mathrm{BP}}^{(c)}\}_{c}\big)$ generalizes multiple BP-level answers across different event categories to infer a personal trait.
}

Note that, along the derivation from records to personality traits, the evidence sets form a nested chain: $\mathcal{E}_{\mathrm{SM}} \subseteq \mathcal{E}_{\mathrm{EM}} \subseteq \mathcal{E}_{\mathrm{BP}} \subseteq \mathcal{E}_{\mathrm{PT}}$.
That is, each higher level introduces both an additional transformation operator and a broader evidence scope.
The hierarchy thus progresses from extracting an explicit fact from one record to inferring a latent disposition from behavior observed across event categories.

\section{The \textsc{Setoka}
  Generation Pipeline} \label{sec:method} Following the hierarchical user understanding framework, we generate user data from abstract to concrete, as illustrated in Fig.~\ref{fig:setoka-pipeline}.
Specifically, the high-level personality traits guide the construction of behavior patterns, which are then mapped to concrete events and records.
We organize this process top-down so that the generated memories remain consistent with the intended personality traits and behavior patterns, while preserving traceable evidence for evaluating each level of user understanding.

\subsection{Personality Trait Generation}
\label{sec:trait-gen}
Prior studies show that personality traits are correlated rather than statistically independent~\cite{vanderlinden2010general}.
For example, extraversion tends to be negatively correlated with neuroticism.
Thus, sampling each trait independently ignores these dependencies and may distort the joint distribution of the resulting personality profiles.

To address this issue, we propose \emph{correlation-aware personality trait sampling}.
Following continuous latent-trait formulations in psychometrics~\cite{haslam2020dimensions,reise2018alternative}, we represent the $d$ trait scores as continuous variables and model their joint distribution with a multivariate Gaussian.
The mean vector $\boldsymbol{\mu}\in\mathbb{R}^{d}$ contains the target population means, and the covariance matrix is \(\boldsymbol{\Sigma}=\mathbf{D}\widetilde{\mathbf{R}}\mathbf{D}\), where $\mathbf{D}$ contains the target standard deviations and $\widetilde{\mathbf{R}}$ is constructed from Pearson correlation coefficients reported in prior studies~\cite{vanderlinden2010general}.
For each synthetic user $u$, we then jointly draw the complete trait vector \(\boldsymbol{\theta}_u \sim \mathcal{N}(\boldsymbol{\mu}, \boldsymbol{\Sigma})\).

By default, we instantiate the framework with the Big Five taxonomy (i.e., $d=5$): openness, conscientiousness, extraversion, agreeableness, and neuroticism~\cite{tupes1992recurrent,soto2017bfi2}.
As in Fig.~\ref{fig:setoka-pipeline}(a), a sampled persona vector may carry a high extraversion score of $0.93$, drawn jointly with the remaining traits so that their combination respects the meta-analytic correlations.
The sampling procedure is taxonomy-agnostic and can be applied to any personality trait inventory for which estimates of trait means, variances, and correlations are available.

\subsection{Behavior Pattern Generation}
\label{sec:bp-gen}
The same personality trait can {lead to} different behavior patterns across event categories~\cite{fleeson2001toward,fleeson2015whole}.
For example, a highly conscientious user may carefully plan work tasks, consistently track personal expenses, and follow a regular exercise routine.
Although these patterns occur in different domains, they can all reflect the same underlying personality trait.
Directly prompting an LLM with a personality trait may instead {generate only a narrow set of stereotypical behaviors}~\cite{cheng2023marked,liu2024evaluating}.

We address this problem with \emph{psychological-scale-based behavior pattern generation}.
A psychological scale is a standardized questionnaire with a predefined scoring rule that converts responses into personality trait scores.
Given a sampled personality trait vector $\boldsymbol{\theta}_u$, we generate a set of questionnaire responses whose scores match $\boldsymbol{\theta}_u$.
Because multiple response combinations can yield the same trait scores, users with the same personality traits can still exhibit different behavior patterns.
We use the 60-item BFI-2~\cite{soto2017bfi2} to represent the Big Five personality traits.
The generated responses are then translated into a behavior pattern $\boldsymbol{\pi}_u$, which guides downstream event generation.
As in Fig.~\ref{fig:setoka-pipeline}(b), an extraversion score of $0.93$ is reflected in item responses such as agreement with ``Is talkative?'', which are translated into concrete patterns such as ``small talk every other day''.

\subsection{Episodic and Semantic Memory Generation}
\label{sec:memory-gen}
Mapping the behavior pattern prior $\boldsymbol{\pi}_u$ into a long-term activity history is difficult to scale with LLM-only generation: as the context grows, earlier behavioral constraints might be forgotten or violated, yielding inconsistent routines or activities that no longer reflect $\boldsymbol{\pi}_u$.
To address this problem, we organize generation as an \emph{event-grounded generation tree} with a coarse-to-fine temporal hierarchy, where higher-level nodes encode behavior patterns and long-term contexts, and lower-level nodes specify concrete events.
{For each parent node, its children are generated through a two-stage process.}
First, the child nodes are generated sequentially, with each node informed by its preceding siblings to preserve experiential continuity.
Second, their details are expanded independently in parallel, avoiding the need to retain the entire event history in context.
This hierarchical and two-stage design bounds the context required at each step while preserving long-range behavioral consistency and local variation.
As in Fig.~\ref{fig:setoka-pipeline}(c), the socializing patterns expand into high-level nodes such as \emph{phone call and follow-up messages} spanning April 20-26, whose child events land on the daily timeline, e.g., \emph{communicated with Alice at home} on 2026-04-20.

To ensure cross-schema consistency, each event serves as a single source of truth from which all corresponding heterogeneous records are derived and linked, as in Fig.~\ref{fig:setoka-pipeline}(d), where the April~20 event yields call, message, and app-usage records sharing the same contact, time window, and location.

\subsection{Query Generation}
\label{sec:query-gen}
Based on the generated user corpus and its generation-time ground truth, we construct level-specific testing instances $(q,a,\mathcal{E})\in\mathcal{T}$ for each of the four levels.
Each query is designed to isolate the additional operator introduced at its level in Eq.~\eqref{eq:hierarchy}.
Because the reference answers and evidence sets are derived from generation-time ground truth, no manual annotation is needed.
An \textbf{SM query} tests direct selection ($\sigma$) from a single record.
We construct it by hiding one field as the answer and using the remaining fields to identify the target record.
An \textbf{EM query} tests whether the system can link ($\lambda$) records that describe the same event.
Following the classical partial-cue-recovery setting~\cite{huet2025episodicmemoriesgenerationevaluation}, the query provides several event cues, such as the time, location, or participants, and asks for a missing detail that must be recovered from the linked records.
A \textbf{BP query} tests aggregation ($\gamma$) over the user's long-term history.
It asks for a behavioral statistic within an activity category, such as how often or under what conditions an activity occurs.
The corresponding statistic computed from the generated event history serves as the reference answer.
A personality trait is latent and is never explicitly stated in any record, so a \textbf{PT query} cannot be evaluated through exact answer matching.
Instead, it asks where a user stands relative to other users on a given trait.
The sampled ground-truth trait vectors define the reference ranking used for evaluation (Sec.~\ref{sec:metrics}).

\subsection{Dataset Statistics}
We use \textsc{Setoka} with DeepSeek-V4-Pro~\cite{deepseekai2026deepseekv4pro} as the generation backbone to create 10 synthetic user profiles.
Each profile contains a five-dimensional personality trait vector, 11 corresponding behavior patterns, and two months of user events.
These events are represented by approximately 1{,}600 heterogeneous records spanning 23 schemas, totaling 364K tokens.
On top of each profile we construct about 1{,}426 queries spanning the four levels.
The factual queries cover all three data models.
The full breakdown is reported in Appendix Table~\ref{tab:dataset-stats}.

\section{Experiments}
\label{sec:experiments}

\begin{table*}[t]
\centering
\small
\setlength{\tabcolsep}{3pt}
\begin{tabular*}{\textwidth}{@{\extracolsep{\fill}} l ccc ccc ccc ccc @{}}
\toprule
& \multicolumn{3}{c}{SM} & \multicolumn{3}{c}{EM} & \multicolumn{3}{c}{BP} & \multicolumn{3}{c}{PT} \\
\cmidrule(lr){2-4}\cmidrule(lr){5-7}\cmidrule(lr){8-10}\cmidrule(lr){11-13}
Memory System & DS & Min & Gma & DS & Min & Gma & DS & Min & Gma & DS & Min & Gma \\
\midrule
Cognee     & \cellcolor{blue!13}28 (46) & \cellcolor{blue!9}20 (32) & \cellcolor{blue!3}6 (71) & \cellcolor{blue!14}31 (91) & \cellcolor{blue!14}32 (100) & \cellcolor{blue!1}2 (90) & \cellcolor{blue!11}\textbf{24 (61)} & \cellcolor{blue!13}\underline{\textbf{28 (86)}} & \cellcolor{blue!1}3 (95) & \cellcolor{blue!9}\textbf{20 (100)} & \cellcolor{blue!11}\underline{\textbf{24 (100)}} & $-8$ (97) \\
HippoRAG~2 & \cellcolor{blue!25}55 (88) & \cellcolor{blue!20}45 (62) & \cellcolor{blue!12}\textbf{26 (97)} & \cellcolor{blue!15}33 (98) & \cellcolor{blue!13}29 (91) & \cellcolor{blue!4}\textbf{9 (99)} & \cellcolor{blue!10}23 (91) & \cellcolor{blue!11}25 (93) & \cellcolor{blue!5}\textbf{12 (100)} & \cellcolor{blue!4}10 (100) & \cellcolor{blue!7}16 (100) & \cellcolor{blue!4}\textbf{9 (98)} \\
Mem0       & \cellcolor{blue!13}29 (71) & \cellcolor{blue!9}20 (37) & \cellcolor{blue!9}19 (84) & \cellcolor{blue!11}25 (89) & \cellcolor{blue!13}28 (98) & \cellcolor{blue!3}6 (89) & \cellcolor{blue!5}11 (51) & \cellcolor{blue!6}13 (55) & \cellcolor{blue!4}10 (85) & 1 (97) & \cellcolor{blue!1}3 (84) & \cellcolor{blue!2}4 (98) \\
MemMachine & \cellcolor{blue!29}65 (88) & \cellcolor{blue!26}58 (74) & \cellcolor{blue!11}25 (92) & \cellcolor{blue!21}\underline{\textbf{46 (97)}} & \cellcolor{blue!19}\textbf{42 (99)} & \cellcolor{blue!4}\textbf{9 (99)} & \cellcolor{blue!10}23 (87) & \cellcolor{blue!13}\underline{\textbf{28 (96)}} & \cellcolor{blue!5}11 (97) & \cellcolor{blue!9}\textbf{20 (100)} & \cellcolor{blue!8}18 (100) & $-8$ (98) \\
Letta      & \cellcolor{blue!23}52 (83) & \cellcolor{blue!19}42 (59) & \cellcolor{blue!9}21 (92) & \cellcolor{blue!14}31 (92) & \cellcolor{blue!14}30 (96) & \cellcolor{blue!3}6 (96) & \cellcolor{blue!10}22 (86) & \cellcolor{blue!11}24 (91) & \cellcolor{blue!3}6 (93) & \cellcolor{blue!4}8 (97) & \cellcolor{blue!8}18 (100) & \cellcolor{blue!1}2 (91) \\
\midrule
DBQuery    & \cellcolor{blue!38}\underline{\textbf{85 (90)}} & \cellcolor{blue!27}\textbf{59 (63)} & \cellcolor{blue!8}17 (73) & \cellcolor{blue!12}27 (90) & \cellcolor{blue!13}28 (99) & 0 (2) & \cellcolor{blue!7}16 (75) & \cellcolor{blue!1}3 (34) & \cellcolor{blue!4}9 (97) & 0 (61) & $-6$ (59) & $-4$ (92) \\
\midrule
Average    & \cellcolor{blue!23}52 (78) & \cellcolor{blue!18}41 (54) & \cellcolor{blue!9}19 (85) & \cellcolor{blue!14}32 (93) & \cellcolor{blue!14}32 (97) & \cellcolor{blue!2}5 (79) & \cellcolor{blue!9}20 (75) & \cellcolor{blue!9}20 (76) & \cellcolor{blue!4}9 (95) & \cellcolor{blue!5}10 (93) & \cellcolor{blue!5}12 (91) & $-1$ (96) \\
Level average & \multicolumn{3}{c}{\cellcolor{blue!17}37 (72)} & \multicolumn{3}{c}{\cellcolor{blue!10}23 (90)} & \multicolumn{3}{c}{\cellcolor{blue!7}16 (82)} & \multicolumn{3}{c}{\cellcolor{blue!3}7 (93)} \\
\bottomrule
\end{tabular*}
\caption{Main results: per-level scores for each memory system $\times$ language model over all 10 users.
Scores are scaled by 100 for display, with the answer rate (\%) in parentheses.
SM, EM, and BP report LLM-judge similarity; PT reports the PT score.
The best score per column is in bold, and the best (system, backbone) cell within each query level is additionally underlined.
The last two rows average each column over the six systems and each level over all systems and backbones.
Model columns: \textbf{DS}~=~DeepSeek-V4-Flash, \textbf{Min}~=~Ministral 3 14B Instruct, \textbf{Gma}~=~Gemma 3 4B-IT.
}
\vspace{-0.2cm}
\label{tab:main-results}
\end{table*}

\subsection{Experimental Setup}
\noindent \textbf{Data Serialization.}
We instantiate the memory construction operator $\mathcal{P}$ as follows.
Following common practice in structured-data understanding~\cite{table-meets-llm,tablebench}, we serialize relational records as Markdown tables due to their high coverage in LLM training corpora and a favorable performance-token trade-off.
Graph-structured data is serialized as adjacency lists, while semi-structured/unstructured records are kept in their native formats.

\noindent \textbf{Backbones and Memory Systems.}
We evaluate 3 language models spanning the deployment spectrum: DeepSeek-V4-Flash~\cite{deepseekai2026deepseekv4flash}, a strong cloud MoE model {with 284B total parameters}; Ministral 3 14B Instruct~\cite{mistralai2026ministral3}, suitable for personal computer deployment; Gemma 3 4B-IT~\cite{gemmateam2025gemma3}, deployable on mobile devices.
We pair each backbone with 5 representative memory systems spanning three  dominant organizational paradigms: Cognee~\cite{markovic2025cognee} and HippoRAG~2~\cite{gutierrez2025hipporag2} are graph-structured systems, Mem0~\cite{chhikara2025mem0} is a flat fact-based system, and MemMachine~\cite{wang2026memmachine} and Letta~\cite{packer2023memgpt} are {hierarchical systems}.

\noindent \textbf{DBQuery Baseline.}
{To examine whether explicit memory construction is necessary, we introduce DBQuery, a memory-free baseline where the LLM directly interacts with the underlying databases.
Given a question, the LLM generates and executes database queries over the raw records and produces an answer, bypassing any intermediate memory representation.
}

\noindent \textbf{Metrics.}
\label{sec:metrics}
We report per-level performance and an overall score averaged across all test instances.
For SM, EM, and BP, we employ an LLM judge (DeepSeek-V4-Pro) to score each answer against the reference on a $[0,1]$ scale, allowing surface-form variation while requiring semantic consistency with the gold standard~\cite{zheng2023llmjudge}.
For PT, we evaluate whether the model correctly preserves the relative ordering of users along each personality trait.
We evaluate the agreement between the predicted and ground-truth user rankings using Kendall's rank correlation coefficient~\cite{kendall1938rank}: \begin{equation}
\mathrm{PT\text{-}Score}_j
=
\tau_{\mathrm{K}}
\left(
\{\hat{\theta}_{u,j}\}_{u \in \mathcal{U}},
\{\theta_{u,j}\}_{u \in \mathcal{U}}
\right),
\end{equation} where $\mathcal{U}$ is the set of users, $\theta_{u,j}$ and $\hat{\theta}_{u,j}$ are the ground-truth and predicted scores of user $u$ on trait dimension $j$, respectively.
The term $\tau_{\mathrm{K}} \in [-1,1]$ denotes Kendall's rank correlation coefficient, where $1$ means the rankings match exactly, $0$ means performance no better than random guessing, and negative values mean opposite rankings.
We report the PT-Score pooled over all $d$ trait dimensions.

\subsection{Main Findings and Discussions}
Table~\ref{tab:main-results} summarizes the main results across memory systems, language models, and levels of user understanding.
We discuss the key findings and their implications below, with additional experimental results provided in Appendix~\ref{sec:appendix}.

\noindent\textbf{Difficulty increases with the level of abstraction.}
As formalized in Eq.~\eqref{eq:hierarchy}, higher levels require both broader evidence integration and additional semantic transformations.
This creates more points at which relevant evidence may be missed, combined incorrectly, or transformed inaccurately.
Accordingly, both the best and average scores decline steadily along the hierarchy.
The best score drops from 0.85 on SM to 0.46 on EM, 0.28 on BP, and 0.24 on PT, while the level average falls from 0.37 to 0.23, 0.16, and 0.07.
\emph{These results show that current memory systems become less effective as queries require broader evidence and more abstract inference, motivating mechanisms for cross-source joining, evidence aggregation, and latent-attribute inference.}

\noindent\textbf{Memory construction helps, but different aggregation levels support different levels of user understanding.}
Memory-based systems outperform DBQuery in all settings except single-record SM lookup, where DBQuery leads with the two stronger backbones DeepSeek and Ministral, but no system performs best at every level.
\ding{182}~DBQuery retrieves individual records without constructing memory.
This preserves exact details, giving the best SM score $0.85$, but performs poorly when multiple records must be combined, dropping to $0.27$ on EM.
\ding{183}~MemMachine is the strongest system overall, ranking first or joint first in five of twelve settings and achieving the best EM score of 0.46.
We attribute this to its comparatively fine-grained record preservation combined with the retrieval of temporally related information, which supplies broader context without excessively compressing the original details.
\ding{184}~Graph-structured memory systems rank first or joint-first on BP and PT in all six settings.
Cognee is best or joint-best with DeepSeek and Ministral (BP $0.24/0.28$, PT $0.20/0.24$), while HippoRAG~2 leads with Gemma (BP $0.12$, PT $0.09$).
Linking related records appears to help aggregate recurring behaviors into higher-level user traits.
But scores vary widely across systems and backbones, so graph structure alone is not enough.
How the graph represents and retrieves evidence matters just as much.
\emph{Overall, separate records preserve precise facts, memories with temporally related context support multi-record reasoning, and explicit associations across records support inference over dispersed evidence.
Memory systems should therefore maintain multiple levels of aggregation and association, and select the level suited to each query.
}

\noindent\textbf{Answer rate reflects calibration rather than competence.}
Answer rate increases from $72\%$ on SM to $93\%$ on PT, even though accuracy decreases.
This is because a system cannot answer a concrete factual question when the relevant evidence is missing, whereas it can always guess an abstract personality trait.
The same contrast appears across backbones.
DeepSeek and Ministral often abstain when evidence is insufficient.
In contrast, the mobile-scale Gemma answers $71$-$100\%$ of queries at every level but achieves the lowest accuracy.
Thus, a high answer rate may indicate unsupported guessing rather than better knowledge.
\emph{Because abstentions are counted as errors in the accuracy metric, answer rate should be reported separately.
Together, the two metrics distinguish systems that abstain when evidence is insufficient from those that answer regardless of uncertainty.
}

\section{Conclusion}
We introduced \textsc{Setoka}, the first benchmark for hierarchical user understanding of memory-augmented personalized agents over heterogeneous data.
\textsc{Setoka} combines a psychology-grounded four-level framework with a psychometrics-based generation pipeline that produces heterogeneous records with ground-truth provenance from the underlying personality traits.
Evaluating 3 language models over 5 memory systems and a SQL-augmented baseline, we find that accuracy decays monotonically with the level of abstraction.
Direct database access largely solves single-record lookup, whereas multi-record assembly and long-horizon aggregation remain unsolved, and even the best configurations rank users by personality trait only marginally above chance.
Improving retrieval alone is therefore insufficient.
\textsc{Setoka} offers a testbed for memory systems that link, aggregate, and generalize user evidence beyond literal recall.

\bibliography{reference/survey,reference/psychology,reference/memory-benchmark,reference/llm-table-understanding,reference/related-work}

\clearpage
\appendix
\counterwithin{figure}{section}
\counterwithin{table}{section}
\counterwithin{algorithm}{section}

\section{Dataset Statistics}
\label{app:dataset-stats}

We release both the generation pipeline proposed in Sec.~\ref{sec:method} and one benchmark instance generated with it.
This instance is used for the memory-system evaluation in the main paper.
The generation pipeline also allows users to generate benchmark instances at other scales by configuring the number of synthetic users and the timeline duration.
Tables~\ref{tab:dataset-stats} and~\ref{tab:schemas} describe the released instance by reporting its dataset statistics and record schemas, respectively.

\begin{table}[h]
\centering
\small
\renewcommand{\arraystretch}{0.95}
\setlength{\tabcolsep}{3pt}

\begin{subtable}{\columnwidth}
\centering
\caption{Corpus}
\label{tab:ds-corpus}
\begin{tabular*}{0.9\columnwidth}{@{\extracolsep{\fill}} l r}
\toprule
User profiles & 10 \\
Schemas & 23 \\
Data models & 3 \\
Records per user & $\sim$1{,}600 \\
Total records & $\sim$16{,}000 \\
Record tokens per user & $\sim$364{,}000 \\
Total record tokens & $\sim$3.64\,M \\
Event span per user & $\sim$2 months \\
\bottomrule
\end{tabular*}
\end{subtable}

\medskip
\begin{subtable}{\columnwidth}
\centering
\caption{Query items}
\label{tab:ds-queries}
\begin{tabular*}{0.9\columnwidth}{@{\extracolsep{\fill}} l r}
\toprule
Factual-QA templates & 868 \\
Factual-QA items & 9{,}898 \\
\quad single-record (SM) & 2{,}375 \\
\quad multi-hop (MH) & 7{,}523 \\
Episodic-reconstruction items (EM) & 4{,}203 \\
Behavior-pattern items (BP) & 110 \\
Personality-trait items (PT) & 50 \\
Total query items & 14{,}261 \\
\bottomrule
\end{tabular*}
\end{subtable}

\medskip
\begin{subtable}{\columnwidth}
\centering
\caption{Factual queries by data-model combination}
\label{tab:ds-modelmix}
\begin{tabular*}{0.9\columnwidth}{@{\extracolsep{\fill}} l r}
\toprule
relational & 4{,}328 \\
document & 500 \\
graph & 398 \\
relational+document & 2{,}319 \\
relational+graph & 424 \\
document+graph & 917 \\
relational+document+graph & 1{,}012 \\
\bottomrule
\end{tabular*}
\end{subtable}

\medskip
\begin{subtable}{\columnwidth}
\centering
\caption{Factual queries by join depth}
\label{tab:ds-depth}
\begin{tabular*}{0.9\columnwidth}{@{\extracolsep{\fill}} l cccc}
\toprule
Join depth & 1 & 2 & 3 & 4 \\
\midrule
Items & 2{,}375 & 1{,}762 & 2{,}922 & 2{,}839 \\
\bottomrule
\end{tabular*}
\end{subtable}

\caption{\textbf{\textsc{Setoka} dataset statistics}.
Query items are counted over all 10 users.
Subtables~(\subref{tab:ds-modelmix}) and~(\subref{tab:ds-depth}) break down the 9{,}898 factual queries by the data models represented in their evidence and by join depth, the number of records combined to answer a query.
}
\label{tab:dataset-stats}
\end{table}

The 23 schemas model the heterogeneous digital traces that a personal assistant may observe across a user's daily life, including 18 relational schemas, 4 document schemas, and 1 graph schema.
The relational schemas record device and file activity, mobility, calls, media consumption, schedules, web activity, payments, and health and exercise.
The document schemas preserve content-rich information from emails, text messages, social posts, and notes, while the graph schema represents the user's contacts and their relationships.
Together, these schemas can describe a daily situation across sources, such as where the user was, whom they communicated with, which digital content they accessed, and what other activities occurred around the same time.

\begin{figure}[!t]
\centering
\includegraphics[width=0.90\linewidth]{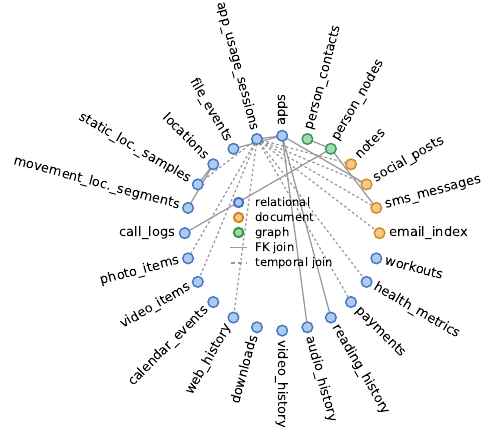}
\caption{\textbf{The schema join graph}.
Nodes are the record schemas of Table~\ref{tab:schemas}, colored by data model.
Solid lines link schemas through matching primary- and foreign-key values.
Dashed lines link a point timestamp to a record whose time interval contains it.
We generate factual-query templates by enumerating simple paths in this graph (Appendix~\ref{sec:query-generation}).
}
\label{fig:schema-graph}
\vspace{\floatsep}
\centering
\includegraphics[width=\linewidth]{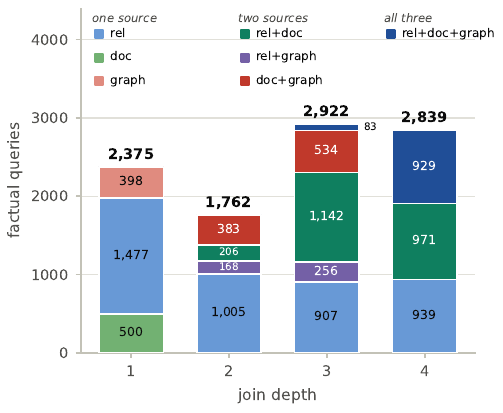}
\caption{Factual queries by join depth, stacked by data-model
combination.}
\label{fig:queries-by-depth}
\end{figure}

\begin{table*}[tp]
\centering
\small
\renewcommand{\arraystretch}{1.18}
\begin{tabular}{@{}l p{0.72\textwidth}@{}}
\toprule
Schema & Fields \\
\midrule
\rowcolor{black!7}\multicolumn{2}{@{}l}{\emph{Relational (18 tables)}} \\
apps & app\_id$^{*}$:str, name:str, install\_time:time, version\_name:str \\
app\_usage\_sessions & session\_id$^{*}$:str, app\_id$^{\dagger}$:str, start\_at:time, end\_at:time \\
file\_events & timestamp$^{*}$:time, file\_name$^{*}$:str, action:str, file\_type:str, file\_path:str, source\_app\_id$^{\dagger}$:str \\
locations & location\_detail$^{*}$:str, location\_name:str \\
static\_location\_samples & sample\_id$^{*}$:str, lat:dec, lon:dec, location\_detail$^{\dagger}$:str, recorded\_at:time \\
movement\_location\_segments & movement\_id$^{*}$:str, start\_lat:dec, start\_lon:dec, end\_lat:dec, end\_lon:dec, start\_location\_detail$^{\dagger}$:str, end\_location\_detail$^{\dagger}$:str, start\_at:time, end\_at:time \\
call\_logs & id$^{*}$:str, contact\_id:str, direction:str, start\_at:time, end\_at:time \\
photo\_items & photo\_id$^{*}$:str, file\_name:str, camera\_model:str, iso:str, album\_name:str, caption:str, ocr\_text:str, timestamp:time \\
video\_items & video\_id$^{*}$:str, file\_name:str, duration\_seconds:int, caption:str, timestamp:time \\
calendar\_events & event\_id$^{*}$:str, title:str, description\_tip:str, start\_at:time, end\_at:time, location:str, created\_at:time \\
web\_history & entry\_id$^{*}$:str, domain:str, url:str, title:str, visited\_at:time \\
downloads & dl\_id$^{*}$:str, file\_name:str, file\_type:str, source\_url:str, downloaded\_at:time \\
video\_history & video\_id$^{*}$:str, video\_name:str, start\_at:time, end\_at:time \\
audio\_history & audio\_id$^{*}$:str, app\_id$^{\dagger}$:str, audio\_type:str, title:str, artist\_or\_channel:str, start\_at:time, end\_at:time \\
reading\_history & reading\_id$^{*}$:str, app\_id$^{\dagger}$:str, reading\_type:str, title:str, author\_or\_source:str, chapter\_or\_section:str, progress:str, start\_at:time, end\_at:time \\
payments & payment\_id$^{*}$:str, amount:dec, payment\_method:str, merchant\_name:str, merchant\_category:str, note\_tip:str, timestamp:time \\
health\_metrics & metric\_id$^{*}$:str, type:str, value:str, unit:str, sample\_at:time \\
workouts & workout\_id$^{*}$:str, type:str, start\_at:time, end\_at:time, distance:int, calories:int \\
\midrule
\rowcolor{black!7}\multicolumn{2}{@{}l}{\emph{Document (4 collections)}} \\
email\_index & email\_id$^{*}$:str, emails:array$\langle$sender:str, subject:str, snippet:str$\rangle$, created\_at:time \\
sms\_messages & msg\_id$^{*}$:str, contact\_id$^{\dagger}$:str, messages:array$\langle$sender:str, content:str$\rangle$, created\_at:time \\
social\_posts & post\_id$^{*}$:str, app\_id$^{\dagger}$:str, messages:array$\langle$author:str, content:str$\rangle$, published\_at:time \\
notes & note\_id$^{*}$:str, title:str, content\_md:str, created\_at:time \\
\midrule
\rowcolor{black!7}\multicolumn{2}{@{}l}{\emph{Graph (nodes and edges)}} \\
person\_nodes (nodes) & contact\_id$^{*}$:str, display\_name:str, relationship:str, phones:array, emails:array \\
person\_contacts (edges) & source\_contact\_id$^{\dagger}$:str, source\_display\_name:str, target\_contact\_id$^{\dagger}$:str, target\_display\_name:str, relationship\_type:str \\
\bottomrule
\end{tabular}
\caption{Record schemas of the released corpus with their fields, grouped
by data model. $^{*}$ marks primary keys and $^{\dagger}$ marks foreign
keys.
Types are abbreviated: str = string, time = timestamp, int = integer, dec = decimal.
Document arrays list their item fields in angle brackets.
Although \texttt{person\_contacts} appears in the table, we treat this edge relation as a join condition rather than as a record schema.
The resulting schema count is therefore 23, as reported in Table~\ref{tab:ds-corpus}.
}
\label{tab:schemas}
\end{table*}

The 23 schemas are connected by 21 join edges.
Of these, 11 match primary- and foreign-key values, and 10 match a point timestamp to a record whose time interval contains it.
Figure~\ref{fig:schema-graph} shows this join graph.
We generate factual-query templates by enumerating simple paths in the graph (Appendix~\ref{sec:query-generation}), and the number of records on each path determines the query's join depth.

The level hierarchy cannot isolate the effect of combining more records.
Across the hierarchy formalized in Eq.~\eqref{eq:hierarchy}, moving to a higher level changes two factors at once: the number of records used and the type of answer requested.
A performance difference between levels may therefore be caused by either factor.
To isolate record-integration difficulty while keeping the answer type factual and the evaluation metric fixed, we introduce \emph{multi-hop} (MH) queries as a natural extension of SM from single-record to multi-record factual retrieval.
An SM query retrieves a fact from one record, whereas an MH query combines 2-4 records across schemas.
We use separate labels for the two settings to make their different record-integration requirements explicit.
Like EM queries, MH queries require linking multiple records; unlike EM queries, they ask for a fact rather than reconstructing an event.
We evaluate MH separately from SM to measure the incremental difficulty of record integration and separately from EM to distinguish factual retrieval from event reconstruction.
Table~\ref{tab:ds-queries} reports the numbers of SM and MH queries, and Fig.~\ref{fig:queries-by-depth} shows their distribution by join depth.
Appendix~\ref{app:heterogeneity} analyzes how accuracy changes with join depth.


\section{Personality Trait Sampling and Verification}
\label{app:trait-sampling}

Section~\ref{sec:trait-gen} presents the basic formulation of our personality trait sampling method.
Here, we provide the details of its implementation.

Let \(N\) denote the number of synthetic users and \(d\) the number of personality trait dimensions.
Prior studies report how personality traits are distributed in a \emph{population}, namely a large group of surveyed human respondents.
The Gaussian proposal is specified by three population-level inputs, each of which can be set using estimates reported in prior studies (Sec.~\ref{sec:trait-gen}): (i)~the vector \(\boldsymbol{\mu}\in\mathbb{R}^{d}\) of population mean scores, one per trait, (ii)~the vector \(\boldsymbol{\sigma}=(\sigma_1,\ldots,\sigma_d)\) of the corresponding standard deviations, and (iii)~the matrix \(\mathbf{R}\in\mathbb{R}^{d\times d}\) of Pearson correlations between every pair of traits.

The procedure runs in three stages.
We first turn the reported statistics into a valid covariance matrix (Sec.~\ref{app:cov-construction}).
We then jointly sample one candidate vector per user, rejecting any candidate that falls outside the valid score ranges (Sec.~\ref{app:bounded-sampling}), and prove that this rejection changes the sampled distribution only slightly (Sec.~\ref{app:rejection-effect}).
We finally verify that the sampled users as a whole are sufficiently close to the target statistics (Sec.~\ref{app:sample-verification}).

\subsection{Converting Reported Statistics into a Valid Covariance Matrix}
\label{app:cov-construction}
Existing studies typically report the standard deviation of each trait and the correlations between pairs of traits, rather than a covariance matrix that can be used directly for joint sampling.
We therefore construct the covariance matrix in two stages.
In Stage 1, we validate the input correlation matrix \(\mathbf{R}\) and repair it if necessary.
In Stage 2, we combine the repaired correlation matrix with the reported standard deviations to obtain the covariance matrix used by our Gaussian sampler.

\paragraph{Stage 1: Validating and Repairing the Correlation Matrix.}
The input correlation matrix \(\mathbf{R}\) may not satisfy the mathematical requirements for Gaussian sampling.
This can happen when its entries are estimated from studies that use different populations, measurement instruments, or analysis methods.
Sampling and rounding errors can introduce further inconsistencies.
A valid correlation matrix must be symmetric, have ones on its diagonal, and be positive semidefinite.
We repair the input matrix by replacing it with the nearest valid correlation matrix in Frobenius norm~\citeapp{higham2002computing}:
\[
\widetilde{\mathbf{R}}
=
\mathop{\arg\min}_{\substack{
\mathbf{X}=\mathbf{X}^{\top},\\
\operatorname{diag}(\mathbf{X})=\mathbf{1},\\
\mathbf{X}\succeq\mathbf{0}
}}
\left\lVert \mathbf{X}-\mathbf{R} \right\rVert_F.
\]
This optimization simultaneously enforces symmetry, unit diagonal entries, and positive semidefiniteness while minimizing the total squared change to the input matrix.

\paragraph{Stage 2: Building the Covariance Matrix.}
Define the diagonal matrix of target standard deviations:
\[
\mathbf{D}
=
\operatorname{diag}(\sigma_1,\ldots,\sigma_d).
\]
Scaling the repaired correlation matrix by these standard deviations gives the target covariance matrix:
\[
\boldsymbol{\Sigma}
=
\mathbf{D}
\widetilde{\mathbf{R}}
\mathbf{D}.
\]
Its diagonal entries are the target variances, \(\Sigma_{jj}=\sigma_j^2\), and each off-diagonal entry is the corresponding covariance, \(\Sigma_{ij}=\sigma_i\widetilde{R}_{ij}\sigma_j\).

Although \(\widetilde{\mathbf{R}}\) is a valid correlation matrix, the resulting \(\boldsymbol{\Sigma}\) may be singular.
We therefore add a small diagonal jitter \(\eta>0\) to obtain a positive-definite sampling covariance:
\[
\boldsymbol{\Sigma}_{\eta}
=
\boldsymbol{\Sigma}+\eta\mathbf{I}.
\]
This adjustment makes \(\boldsymbol{\Sigma}_{\eta}\) positive definite while leaving its off-diagonal covariances unchanged.

\subsection{Sampling Correlated Trait Vectors within Valid Ranges}
\label{app:bounded-sampling}
For each user \(u\), we sample a correlated personality trait vector directly from
\[
\boldsymbol{\theta}_u
\sim
\mathcal{N}(\boldsymbol{\mu},\boldsymbol{\Sigma}_{\eta}).
\]
Each sampled trait score must later be realized by a questionnaire answer sheet (Sec.~\ref{sec:bp-gen}).
Because questionnaire responses are bounded and each trait score is computed from those responses, trait \(j\) can attain scores only within a range \([l_j,r_j]\).
A Gaussian distribution is unbounded and may produce scores outside these attainable ranges.
We therefore accept a sampled vector only if every dimension $j$ satisfies
\[
l_j \leq \theta_{u,j} \leq r_j,
\qquad
\forall j\in\{1,\ldots,d\}.
\]
For each accepted vector, Appendix~\ref{app:scale-mapping} constructs an answer sheet whose scores match the sampled targets within a prescribed tolerance.
In our BFI-2 setting, trait scores are normalized so that
\([l_j,r_j]=[0,1]\) for every dimension.
If any dimension falls outside its range, the entire vector is rejected and resampled.
A natural alternative would clip each out-of-range value to the nearest boundary.
Clipping, however, causes many sampled scores to equal the boundary values and can distort the correlations between traits.
Rejecting the whole vector avoids both problems.
The next subsection measures how rejection changes the sampled distribution.

\subsection{Effect of Rejection Sampling on the Resulting Distribution}
\label{app:rejection-effect}

Rejection sampling restricts the Gaussian to the multidimensional interval (box) \(\mathcal{B}=\prod_{j=1}^{d}[l_j,r_j]\), so the accepted vectors follow a truncated distribution rather than \(\mathcal{N}(\boldsymbol{\mu},\boldsymbol{\Sigma}_{\eta})\) itself.
Informally, if only a small fraction of candidate vectors is rejected, then keeping only the accepted ones changes the distribution only slightly.
The size of this change depends on the rejection probability.

Let \(\delta=\Pr[\boldsymbol{\theta}\notin\mathcal{B}]\) for \(\boldsymbol{\theta}\sim\mathcal{N}(\boldsymbol{\mu},\boldsymbol{\Sigma}_{\eta})\).
Let \(\boldsymbol{\theta}^{\mathcal{B}}\) denote the accepted vector, and write \(s_j=\sqrt{(\boldsymbol{\Sigma}_{\eta})_{jj}}\).
In words, \(\delta\) is the rejection probability, and \(s_j\) is the standard deviation of trait \(j\) before rejection.
The goal of the following analysis is to determine how enforcing valid score ranges affects both the sampled population and evaluations based on it.
The rejection probability \(\delta\) provides a common quantity for controlling these effects.
We first prove that conditioning on acceptance changes any pre-rejection distribution by exactly \(\delta\) in total variation distance (Proposition~1).
We then use this distribution-wide result to bound the corresponding change in the population Kendall \(\tau_a\) used for evaluation (Corollary~1).
Because benchmark generation also depends on preserving the target trait statistics, we next prove bounds on changes in trait means and target-centered second moments (Proposition~2).
These moment bounds yield guarantees for covariance and pairwise correlation (Corollary~2).
Finally, we bound \(\delta\) using the marginal Gaussian tail probabilities of our sampler.
Substituting this bound into the preceding results gives explicit numerical guarantees for the sampler used in our benchmark.

Total variation distance, used in Proposition~1, is a standard probability metric that gives the largest probability difference over all measurable sets~\citeapp{gibbs2002choosing}.
It provides a distribution-wide guarantee: rejection changes the probability of any set of personality profiles by at most \(\delta\).

\textbf{Proposition 1} (Distributional effect of rejection)\textbf{.}
\emph{Let \(\mathbf{X}\) follow any probability distribution on \(\mathbb{R}^{d}\), and let \(\mathcal{B}\subseteq\mathbb{R}^{d}\) be a measurable acceptance region with \(\Pr[\mathbf{X}\in\mathcal{B}]=1-\delta>0\).
Let \(\mathbf{X}^{\mathcal{B}}\) follow the conditional distribution of \(\mathbf{X}\) given \(\mathbf{X}\in\mathcal{B}\).
Then}
\[
\begin{aligned}
d_{\mathrm{TV}}(\mathbf{X}^{\mathcal{B}},\mathbf{X})
&:=
\sup_{A\subseteq\mathbb{R}^{d}}
\bigl|\Pr[\mathbf{X}^{\mathcal{B}}\in A] \\
&\qquad
-\Pr[\mathbf{X}\in A]\bigr|
=\delta.
\end{aligned}
\]

\textbf{Proof.}
Set \(p=1-\delta\).
For any measurable set \(A\), write \(a=\Pr[\mathbf{X}\in A\cap\mathcal{B}]\) and \(b=\Pr[\mathbf{X}\in A\cap\mathcal{B}^{c}]\).
Because \(\mathbf{X}^{\mathcal{B}}\) follows the conditional distribution,
\begin{align*}
\Pr[\mathbf{X}^{\mathcal{B}}\in A]
-\Pr[\mathbf{X}\in A]
&=
\frac{a}{p}-(a+b) \\
&=
\frac{\delta a}{p}-b.
\end{align*}
Because \(0\leq \delta a/p\leq\delta\) and \(0\leq b\leq\delta\), the absolute difference is at most \(\delta\) for every \(A\).
Taking \(A=\mathcal{B}^{c}\) attains this bound:
\[
\bigl|\Pr[\mathbf{X}^{\mathcal{B}}\in\mathcal{B}^{c}]
-\Pr[\mathbf{X}\in\mathcal{B}^{c}]\bigr|
=
|0-\delta|
=
\delta.
\tag*{\(\square\)}
\]

\begin{figure}[!t]
\centering
\includegraphics[width=0.92\linewidth]{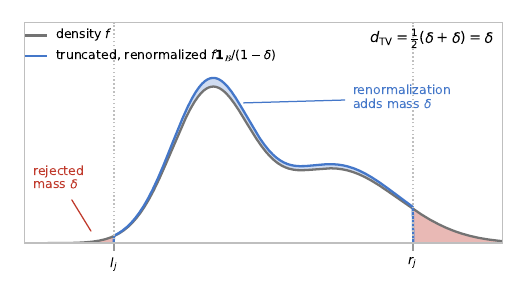}
\caption{The continuous case of Proposition~1 for a generic density \(f\), restricted to one coordinate of the valid box \(\mathcal{B}\).
Rejection removes the mass \(\delta\) outside the valid range and renormalization adds the same mass inside, so the total variation distance equals \(\delta\).
}
\label{fig:prop1-tv}
\end{figure}

Figure~\ref{fig:prop1-tv} illustrates this identity.

\medskip
\noindent\textbf{Corollary 1} (Population-level Kendall \(\tau_a\) stability)\textbf{.}
\emph{Fix a model \(m\) and trait dimension \(j\), and let \(P_0\) and \(P_{\mathcal{B}}\) denote the trait-vector distributions before and after rejection, respectively.
Assume that the same downstream generation and evaluation process maps each trait vector to the predicted score \(\hat{\theta}_{m,j}\) and is defined on the supports of both distributions.
For trait vectors outside \(\mathcal{B}\), this common process is a hypothetical extension used only to define the population comparison with \(P_0\).
Let \(\tau^{(a)}_{m,j}(P)\) be the pairwise population Kendall \(\tau_a\) between \(\theta_j\) and \(\hat{\theta}_{m,j}\) when trait vectors follow \(P\).
Tied pairs contribute zero.
Then}
\[
\begin{aligned}
\bigl|\tau^{(a)}_{m,j}(P_{\mathcal{B}})-\tau^{(a)}_{m,j}(P_0)\bigr|
&\leq
2\left[1-(1-\delta)^2\right] \\
&=
4\delta-2\delta^2.
\end{aligned}
\]

\textbf{Proof.}
Let \(Q_{m,j,0}\) and \(Q_{m,j,\mathcal{B}}\) denote the induced single-user distributions of \(z=(\theta_j,\hat{\theta}_{m,j})\) under \(P_0\) and \(P_{\mathcal{B}}\), respectively.
Because both distributions pass through the same downstream process, the data-processing property of total variation and Proposition~1 give
\[
d_{\mathrm{TV}}(Q_{m,j,\mathcal{B}},Q_{m,j,0})
\leq
d_{\mathrm{TV}}(P_{\mathcal{B}},P_0)
=
\delta.
\]
Write \(\varepsilon=d_{\mathrm{TV}}(Q_{m,j,\mathcal{B}},Q_{m,j,0})\).
By the maximal-coupling characterization of total variation~\citeapp{gibbs2002choosing}, the two single-user outcomes can be coupled to match with probability \(1-\varepsilon\).
Coupling two users independently makes both outcomes match with probability \((1-\varepsilon)^2\), so
\[
\begin{aligned}
d_{\mathrm{TV}}
\left(
Q_{m,j,\mathcal{B}}^{\otimes 2},
Q_{m,j,0}^{\otimes 2}
\right)
&\leq
1-(1-\varepsilon)^2 \\
&\leq
1-(1-\delta)^2.
\end{aligned}
\]
Here, \(Q^{\otimes 2}\) denotes the joint distribution of two independent users drawn from \(Q\).
For \(z=(\theta_j,\hat{\theta}_{m,j})\) and \(z'=(\theta'_j,\hat{\theta}'_{m,j})\), the Kendall \(\tau_a\) kernel is
\[
h_{m,j}(z,z')
=
\operatorname{sign}
\left[
(\theta_j-\theta'_j)
(\hat{\theta}_{m,j}-\hat{\theta}'_{m,j})
\right]
\in[-1,1],
\]
where tied pairs contribute zero.
The two population correlations are the corresponding kernel expectations:
\[
\begin{aligned}
\tau^{(a)}_{m,j}(P_{\mathcal{B}})
&=
\mathbb{E}_{Q_{m,j,\mathcal{B}}^{\otimes 2}}
\left[h_{m,j}\right], \\
\tau^{(a)}_{m,j}(P_0)
&=
\mathbb{E}_{Q_{m,j,0}^{\otimes 2}}
\left[h_{m,j}\right].
\end{aligned}
\]
For any distributions \(R\) and \(S\) and any bounded function \(h\),
\[
\left|
\mathbb{E}_{R}[h]-\mathbb{E}_{S}[h]
\right|
\leq
2\lVert h\rVert_{\infty}
d_{\mathrm{TV}}(R,S).
\]
Since \(\lVert h_{m,j}\rVert_{\infty}\leq 1\), applying this inequality to the two product distributions gives
\begin{align*}
\left|
\tau^{(a)}_{m,j}(P_{\mathcal{B}})
-\tau^{(a)}_{m,j}(P_0)
\right|
&\leq
2d_{\mathrm{TV}}
\left(
Q_{m,j,\mathcal{B}}^{\otimes 2},
Q_{m,j,0}^{\otimes 2}
\right) \\
&\leq
2\left[1-(1-\delta)^2\right] \\
&=
4\delta-2\delta^2.
\tag*{\(\square\)}
\end{align*}

This is a population-level result.
It does not include finite-sample variation or the dependence introduced by the group-level sample verification of Appendix~\ref{app:sample-verification}.

Total variation controls probability differences over sets, but it does not directly bound the unbounded first and second moments.
The next proposition does not require Gaussianity; it assumes only finite marginal fourth moments and gives separate bounds for the mean and the second moment centered at the target mean \(\boldsymbol{\mu}\).

\textbf{Proposition 2} (Effect of rejection on the first and second moments)\textbf{.}
\emph{Assume that the pre-rejection vector \(\boldsymbol{\theta}\) has mean \(\boldsymbol{\mu}\), covariance \(\boldsymbol{\Sigma}_{\eta}\), and marginal standard deviations \(s_j>0\).
Define its standardized fourth moments by}
\[
\kappa_j
=
\frac{\mathbb{E}[(\theta_j-\mu_j)^4]}{s_j^4}
<
\infty.
\]
\emph{Then:}
\begingroup
\setlength{\leftmargini}{1.8em}
\begin{enumerate}
\setlength{\itemsep}{2pt}
\setlength{\parsep}{0pt}
\setlength{\topsep}{2pt}
\item[\emph{(i)}] \emph{For every \(j\),}
      \[
      \bigl|\mathbb{E}[\theta^{\mathcal{B}}_j]-\mu_j\bigr|
      \leq \frac{s_j\sqrt{\delta}}{1-\delta}.
      \]
\item[\emph{(ii)}] \emph{For every \(i,j\),}
      \[
      \begin{aligned}
&\bigl|\mathbb{E}[(\theta^{\mathcal{B}}_i-\mu_i)
(\theta^{\mathcal{B}}_j-\mu_j)]
-(\boldsymbol{\Sigma}_{\eta})_{ij}\bigr| \\
&\qquad\leq
\frac{s_i s_j\bigl[\delta+(\kappa_i\kappa_j)^{1/4}\sqrt{\delta}\bigr]}{1-\delta}.
\end{aligned}
      \]
\end{enumerate}
\endgroup
\noindent\emph{Hence every first and target-centered second moment of the accepted distribution differs from its target by \(O(\sqrt{\delta})\).}

\textbf{Proof.}
The accepted distribution is obtained by discarding the rejected candidates, whose total probability is \(\delta\).
Any change in an expectation must therefore come from those rejected candidates.
Write \(X_j=\theta_j-\mu_j\), \(X^{\mathcal{B}}_j=\theta^{\mathcal{B}}_j-\mu_j\), and \(p=1-\delta\).
Let \(\mathbf{1}_{\mathcal{B}}\) equal 1 when a vector lies in \(\mathcal{B}\) and 0 otherwise.

(i) Since \(\mathbb{E}[X_j]=0\), we have
\[
\mathbb{E}[X_j\mathbf{1}_{\mathcal{B}}]
=-\mathbb{E}[X_j\mathbf{1}_{\mathcal{B}^{c}}],
\]
and therefore
\[
\mathbb{E}[\theta^{\mathcal{B}}_j]-\mu_j
=-\frac{\mathbb{E}[X_j\mathbf{1}_{\mathcal{B}^{c}}]}{p}.
\]
By the Cauchy-Schwarz inequality,
\[
\bigl|\mathbb{E}[X_j\mathbf{1}_{\mathcal{B}^{c}}]\bigr|
\leq
\sqrt{\mathbb{E}[X_j^{2}]\,\Pr[\mathcal{B}^{c}]}
=s_j\sqrt{\delta}.
\]

(ii) Splitting the expectation into accepted and rejected regions in the
same way gives
\[
\mathbb{E}[X^{\mathcal{B}}_iX^{\mathcal{B}}_j]
-(\boldsymbol{\Sigma}_{\eta})_{ij}
=\frac{\delta(\boldsymbol{\Sigma}_{\eta})_{ij}
-\mathbb{E}[X_iX_j\mathbf{1}_{\mathcal{B}^{c}}]}{p}.
\]
By the definition of \(\kappa_j\),
\[
\mathbb{E}[X_j^{4}]
=
\kappa_j s_j^{4}.
\]
The Cauchy-Schwarz inequality gives
\[
\mathbb{E}[X_i^{2}X_j^{2}]
\leq
\sqrt{\mathbb{E}[X_i^{4}]\,\mathbb{E}[X_j^{4}]}
=
\sqrt{\kappa_i\kappa_j}\,s_i^{2}s_j^{2}.
\]
Applying it once more gives
\[
\bigl|\mathbb{E}[X_iX_j\mathbf{1}_{\mathcal{B}^{c}}]\bigr|
\leq
(\kappa_i\kappa_j)^{1/4}s_is_j\sqrt{\delta}.
\]
Combining the two terms with \(|(\boldsymbol{\Sigma}_{\eta})_{ij}|\leq s_is_j\) proves the claim.
\hfill\(\square\)

For the Gaussian distribution used by our sampler, \(\kappa_j=3\) for every \(j\), so \((\kappa_i\kappa_j)^{1/4}=\sqrt{3}\).

Proposition~2(ii) bounds the second moment taken around the target mean \(\boldsymbol{\mu}\), whereas the covariance of the accepted vectors is centered around their own mean.
Because rejection can shift that mean, an additional correction term is required.
The next corollary combines Proposition~2(i) and Proposition~2(ii) to convert the second-moment bound into a bound on the covariance itself.

\textbf{Corollary 2} (Covariance and correlation preservation after rejection)\textbf{.}
\emph{Under the assumptions of Proposition~2, for every \(i,j\),}
\[
\begin{aligned}
&\bigl|\operatorname{Cov}(\theta^{\mathcal{B}}_i,\theta^{\mathcal{B}}_j)
-(\boldsymbol{\Sigma}_{\eta})_{ij}\bigr| \\
&\qquad\leq
s_is_j
\left[
\frac{\delta+(\kappa_i\kappa_j)^{1/4}\sqrt{\delta}}{1-\delta}
+
\frac{\delta}{(1-\delta)^{2}}
\right].
\end{aligned}
\]
\emph{Thus, the covariance error is \(O(\sqrt{\delta})\).
If the target variances are positive, the pairwise-correlation error is also \(O(\sqrt{\delta})\) for sufficiently small \(\delta\).}

\textbf{Proof.}
Write the covariance as \(\operatorname{Cov}(\theta^{\mathcal{B}}_i,\theta^{\mathcal{B}}_j) =\mathbb{E}[X^{\mathcal{B}}_iX^{\mathcal{B}}_j] -\mathbb{E}[X^{\mathcal{B}}_i]\,\mathbb{E}[X^{\mathcal{B}}_j]\).
The first term deviates from \((\boldsymbol{\Sigma}_{\eta})_{ij}\) by at most the bound of Proposition~2(ii).
The second term is bounded using Proposition~2(i) as \(\bigl|\mathbb{E}[X^{\mathcal{B}}_i]\bigr| \bigl|\mathbb{E}[X^{\mathcal{B}}_j]\bigr| \leq s_is_j\,\delta/(1-\delta)^{2}\).
Adding these two bounds proves the covariance result.
Setting \(i=j\) gives an \(O(\sqrt{\delta})\) variance error.
Because the target variances are positive, the accepted variances remain bounded away from zero for sufficiently small \(\delta\).
Normalizing the covariances by the corresponding standard deviations therefore gives the stated pairwise-correlation bound.
\hfill\(\square\)

In our setting, every valid range is symmetric around its mean.
The reflection \(\boldsymbol{\theta}\mapsto 2\boldsymbol{\mu}-\boldsymbol{\theta}\) maps accepted vectors to accepted vectors.
It also leaves the Gaussian density unchanged.
Consequently, every accepted vector and its reflection occur with the same density, so the accepted distribution is symmetric around \(\boldsymbol{\mu}\).
A distribution that is symmetric around a point has that point as its mean.
The accepted mean therefore equals \(\boldsymbol{\mu}\) exactly, and the covariance then equals the second moment bounded in Proposition~2(ii).
Substituting \((\kappa_i\kappa_j)^{1/4}=\sqrt{3}\) into that bound gives, for every \(i,j\),
\[
\left|
\operatorname{Cov}(\theta_i^{\mathcal{B}},\theta_j^{\mathcal{B}})
-(\boldsymbol{\Sigma}_{\eta})_{ij}
\right|
\leq
s_is_j\frac{\delta+\sqrt{3\delta}}{1-\delta}.
\]
This bound is tighter than the general bound in Corollary~2 because symmetry eliminates the mean-shift term.

\begin{figure}[!t]
\centering
\includegraphics[width=0.92\linewidth]{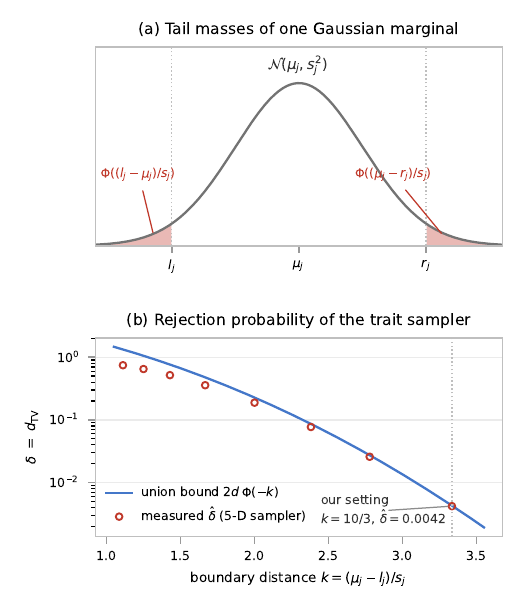}
\caption{\textbf{Bounding the rejection probability of the Gaussian trait sampler}.
(a)~One Gaussian marginal, drawn with \(s_j=0.25\) so that the tails are visible.
The two shaded tail masses are the per-dimension terms of the union bound.
(b)~Measured rejection probability of the five-dimensional sampler as the valid ranges move away from the means, using 2-8 million candidate vectors per point, compared with the union bound.
}
\label{fig:rejection-bound}
\end{figure}

\paragraph{Bounding the rejection probability.}
To connect the preceding results to our Gaussian sampler, we now bound the probability that a candidate vector is rejected.
The probability that any dimension leaves its range is at most the sum of the per-dimension probabilities.
This union bound gives \(\delta\leq\sum_{j=1}^{d}\bigl[\Phi\bigl((l_j-\mu_j)/s_j\bigr) +\Phi\bigl((\mu_j-r_j)/s_j\bigr)\bigr]\), where \(\Phi\) is the standard Gaussian CDF.
The two terms for dimension \(j\) are the probabilities of \(\theta_j\) falling below \(l_j\) or above \(r_j\), shown in Fig.~\ref{fig:rejection-bound}(a).
In our sampler every trait dimension uses the same margins, with \(\mu_j=1/2\), \(s_j=0.15\), and \([l_j,r_j]=[0,1]\), so all \(2d\) tail terms are equal and the bound reduces to
\[
\delta
\leq
2d\,\Phi(-k),
\qquad
k=\frac{\mu_j-l_j}{s_j}=\frac{10}{3}.
\]
The Gaussian tail bound \(\Phi(-k)\leq e^{-k^{2}/2}/\bigl(k\sqrt{2\pi}\bigr)\) makes the exponential decay in \(k\) explicit.
Evaluating the reduced bound at \(k=10/3\) with \(d=5\) gives \(\delta\leq 10\,\Phi(-10/3)\approx 4.3\times10^{-3}\).
Figure~\ref{fig:rejection-bound}(b) compares this bound with the rejection probability measured from simulations of the exact sampling procedure, using the correlation matrix built from the meta-analytic Big Five correlations~\cite{vanderlinden2010general}.
The measured rejection probability in our setting is \(0.0042\), close to the union bound of \(0.0043\).
Using \(\kappa_j=3\) for the Gaussian distribution and substituting the theoretical union bound \(\delta\leq0.0043\) into the preceding results gives the following guarantees.
\begin{table}[H]
\centering
\small
\begin{tabular}{@{}>{\raggedright\arraybackslash}p{0.24\linewidth}>{\raggedright\arraybackslash}p{0.39\linewidth}>{\raggedright\arraybackslash}p{0.25\linewidth}@{}}
\toprule
Result & Quantity & Bound \\
\midrule
Proposition~1 & Set-probability change & \(0.0043\) \\
Corollary~1 & Population Kendall \(\tau_a\) change & \(0.018\) \\
Proposition~2(i) & Mean shift & \(0.066s_j\) \\
Proposition~2(ii) & Target-centered second-moment error & \(0.119s_is_j\) \\
Corollary~2 & Covariance error & \(0.123s_is_j\) \\
Corollary~2 & Pairwise-correlation error & \(O(\sqrt{\delta})\) \\
Symmetric setting & Mean shift; covariance error & \(0;\ 0.119s_is_j\) \\
\bottomrule
\end{tabular}
\caption{Numerical implications of the rejection-probability bound in our Gaussian sampler.}
\label{tab:rejection-implications}
\end{table}
The symmetric-setting covariance bound is tighter than the general Corollary~2 bound because the accepted mean equals the target mean exactly.

The preceding results bound only the systematic bias introduced by rejection.
Statistics computed from a finite sample of \(N\) users also fluctuate around these moments.
The verification below therefore measures the empirical statistics directly and regenerates the sample whenever any deviation exceeds its tolerance.

\begin{algorithm}[t]
\caption{Correlation-Aware Personality Trait Sampling and Verification}
\label{alg:trait-sampling}
\begin{algorithmic}[1]
\Require
Number of users \(N\);
target population means \(\boldsymbol{\mu}\);
target standard deviations \(\boldsymbol{\sigma}\);
correlation matrix \(\mathbf{R}\);
personality trait bounds \(\{[l_j,r_j]\}_{j=1}^{d}\);
tolerances \(\epsilon_{\mu},\epsilon_{\sigma},\epsilon_R\);
maximum user-level retries \(K_{\mathrm{user}}\);
maximum sample-level retries \(K_{\mathrm{samp}}\);
numerical jitter \(\eta\)

\Ensure
Verified personality trait matrix
\(\mathbf{T}\in\mathbb{R}^{N\times d}\)

\State
Set \(\widetilde{\mathbf{R}}\) to the valid correlation matrix nearest to \(\mathbf{R}\) in Frobenius norm

\State
\(\mathbf{D}
\gets
\operatorname{diag}(\sigma_1,\ldots,\sigma_d)\)

\State
\(\boldsymbol{\Sigma}
\gets
\mathbf{D}\widetilde{\mathbf{R}}\mathbf{D}\)

\State
\(\boldsymbol{\Sigma}_{\eta}
\gets
\boldsymbol{\Sigma}+\eta\mathbf{I}\)

\For{\(a=1,\ldots,K_{\mathrm{samp}}\)}
    \State
    \(\mathbf{T}\gets\mathbf{0}_{N\times d}\)
    \State
    \(\mathrm{sampleValid}\gets\mathrm{true}\)

    \For{\(u=1,\ldots,N\)}
        \State
        \(\mathrm{userAccepted}\gets\mathrm{false}\)

        \For{\(b=1,\ldots,K_{\mathrm{user}}\)}
            \State
            Sample
            \(\boldsymbol{\theta}\sim
            \mathcal{N}(\boldsymbol{\mu},\boldsymbol{\Sigma}_{\eta})\)

            \If{
            \(l_j\leq\theta_j\leq r_j\)
            for every \(j=1,\ldots,d\)
            }
                \State
                \(\mathbf{T}_{u,:}
                \gets
                \boldsymbol{\theta}^{\top}\)

                \State
                \(\mathrm{userAccepted}
                \gets
                \mathrm{true}\)

                \State \textbf{break}
            \EndIf
        \EndFor

        \If{\(\mathrm{userAccepted}=\mathrm{false}\)}
            \State
            \(\mathrm{sampleValid}
            \gets
            \mathrm{false}\)

            \State \textbf{break}
        \EndIf
    \EndFor

    \If{\(\mathrm{sampleValid}=\mathrm{false}\)}
        \State \textbf{continue}
    \EndIf

    \State
    Compute empirical mean
    \(\widehat{\boldsymbol{\mu}}\)
    from \(\mathbf{T}\)

    \State
    Compute empirical standard deviations
    \(\widehat{\boldsymbol{\sigma}}\)
    from \(\mathbf{T}\)

    \State
    Compute empirical correlation matrix
    \(\widehat{\mathbf{R}}\)
    from \(\mathbf{T}\)

    \State
    \(e_{\mu}
    \gets
    \displaystyle
    \max_j
    \frac{|\widehat{\mu}_j-\mu_j|}{\sigma_j}\)

    \State
    \(e_{\sigma}
    \gets
    \displaystyle
    \max_j
    \frac{|\widehat{\sigma}_j-\sigma_j|}{\sigma_j}\)

    \State
    \(e_R
    \gets
    \displaystyle
    \max_{i<j}
    |\widehat{R}_{ij}
    -
    \widetilde{R}_{ij}|\)

    \If{
    \(e_{\mu}\leq\epsilon_{\mu}\)
    \textbf{and}
    \(e_{\sigma}\leq\epsilon_{\sigma}\)
    \textbf{and}
    \(e_R\leq\epsilon_R\)
    }
        \State \Return \(\mathbf{T}\)
    \EndIf
\EndFor

\State
\textbf{raise} sampling failure after
\(K_{\mathrm{samp}}\) attempts
\end{algorithmic}
\end{algorithm}

\subsection{Verifying the Sampled Users Against the Target Statistics}
\label{app:sample-verification}
Both per-user rejection and the small number of users can move the sampled group away from the target statistics.
This stage therefore checks the group as a whole and regenerates it when the deviation is too large.
After sampling all \(N\) users, we arrange their personality trait vectors as \[ \mathbf{T} = \begin{bmatrix}
\boldsymbol{\theta}_1^{\top}\\
\vdots\\
\boldsymbol{\theta}_N^{\top}
\end{bmatrix} \in \mathbb{R}^{N\times d}.
\]
We compute the empirical mean and standard deviation of each
dimension:
\[
\widehat{\mu}_j
=
\frac{1}{N}
\sum_{u=1}^{N}
\theta_{u,j},
\]
\[
\widehat{\sigma}_j
=
\sqrt{
\frac{1}{N-1}
\sum_{u=1}^{N}
\left(
\theta_{u,j}-\widehat{\mu}_j
\right)^2
}.
\]
The empirical correlation between dimensions \(i\) and \(j\) is
\[
\widehat{R}_{ij}
=
\frac{
\sum_{u=1}^{N}
(\theta_{u,i}-\widehat{\mu}_i)
(\theta_{u,j}-\widehat{\mu}_j)
}{
(N-1)\widehat{\sigma}_i\widehat{\sigma}_j
}.
\]

We use the following normalized errors to measure deviations from the target population statistics.
\[ e_{\mu} = \max_j \frac{ \left| \widehat{\mu}_j-\mu_j \right| }{ \sigma_j }, \]
\[ e_{\sigma} = \max_j \frac{ \left| \widehat{\sigma}_j-\sigma_j \right| }{ \sigma_j }, \]
\[ e_R = \max_{i<j} \left| \widehat{R}_{ij} - \widetilde{R}_{ij} \right|.
\]
The quantities \(e_{\mu}\) and \(e_{\sigma}\) are the largest mean and
standard-deviation errors across dimensions, normalized by the target
standard deviations.
The quantity \(e_R\) is the largest absolute error across all pairwise correlations.
The sample of \(N\) users is accepted if \[ e_{\mu}\leq\epsilon_{\mu}, \qquad e_{\sigma}\leq\epsilon_{\sigma}, \qquad e_R\leq\epsilon_R, \] where \(\epsilon_{\mu}\), \(\epsilon_{\sigma}\), and \(\epsilon_R\) are predefined tolerances.
If any condition is violated, the entire sample is regenerated.
We impose a maximum number of retries at the user and sample levels to prevent an unbounded rejection loop.

Algorithm~\ref{alg:trait-sampling} summarizes the complete procedure.

\section{Item-Level Response Profile Synthesis}
\label{app:scale-mapping}
This section formalizes the \emph{psychological-scale-based behavior pattern generation} of Sec.~\ref{sec:bp-gen}: given a sampled personality trait vector \(\boldsymbol{\theta}_u\), we synthesize an item-level response profile whose scale scores are consistent with \(\boldsymbol{\theta}_u\), while allowing different response configurations to realize the same personality trait scores.
\subsection{Scale Structure and Answer-Sheet Synthesis}
The Big Five Inventory-2 (BFI-2) is a psychological questionnaire that measures five broad personality domains and their more specific facets~\cite{soto2017bfi2}.
It contains 60 personality statements rated from 1 (disagree strongly) to 5 (agree strongly).
The statements are grouped into five domains, with three facets per domain and four items per facet; extraversion, for example, includes sociability, assertiveness, and energy level.
Suppose a respondent selects 5 for an item about being talkative but 2 for an item about being reserved.
Reserved behavior points in the opposite direction from extraversion, so the second item is reverse-keyed: its response is converted from 2 to 4, and both answers then indicate higher extraversion.
Averaging the aligned answers produces facet and domain scores.
Standard use maps item responses to trait scores; our pipeline starts from target trait scores and constructs one plausible answer sheet that yields them.

We formalize this scale structure and scoring process as follows.
Consider a psychological scale with \(m\) items, each rated on an \(L\)-point Likert scale.
The items are partitioned into \(d\) domains, one per personality trait dimension, through index sets \(\mathcal{I}_1,\ldots,\mathcal{I}_d\subseteq\{1,\ldots,m\}\), and each domain \(j\) is further partitioned into facets \(\mathcal{F}_{j,1},\ldots,\mathcal{F}_{j,c_j}\).
Each item \(i\) carries a keying direction \(k_i\in\{+1,-1\}\), where \(k_i=-1\) marks reverse-keyed items.
For the 60-item BFI-2 version we use~\cite{soto2017bfi2}, \(m=60\), \(L=5\), \(d=5\), \(|\mathcal{I}_j|=12\), and \(c_j=3\) with four items per facet.

We represent the responses of user \(u\) as an answer sheet, \[ \mathbf{r}_u = (r_{u,1},\ldots,r_{u,m}) \in \{1,\ldots,L\}^{m}.
\]
The scoring rule first aligns each response with its keying direction,
\[
\tilde{r}_{u,i}
=
\begin{cases}
r_{u,i}, & k_i=+1,\\
L+1-r_{u,i}, & k_i=-1,
\end{cases}
\]
and then averages the aligned responses within each facet and domain:
\[
s_{u,j,f}
=
\frac{1}{|\mathcal{F}_{j,f}|}
\sum_{i\in\mathcal{F}_{j,f}}
\tilde{r}_{u,i},
\qquad
s_{u,j}
=
\frac{1}{|\mathcal{I}_j|}
\sum_{i\in\mathcal{I}_j}
\tilde{r}_{u,i}.
\]
We write
\(\mathbf{s}(\mathbf{r}_u)=(s_{u,1},\ldots,s_{u,d})\)
for the domain-score vector induced by an answer sheet.
Because sampled personality trait scores live on the normalized \([0,1]\) scale of Appendix~\ref{app:trait-sampling}, we map each raw domain score from \([1,L]\) onto the same scale through \[ \bar{s}_{u,j} = \frac{s_{u,j}-1}{L-1}, \] and write \(\bar{\mathbf{s}}(\mathbf{r}_u)=(\bar{s}_{u,1},\ldots,\bar{s}_{u,d})\) for the normalized domain-score vector.

\paragraph{Consistency constraint.}
An answer sheet is consistent with the sampled personality trait vector if every recovered normalized domain score falls within a tolerance \(\epsilon\) of the corresponding target score.
The set of consistent answer sheets is \[ \mathcal{R}(\boldsymbol{\theta}_u) = \left\{ \mathbf{r}\in\{1,\ldots,L\}^{m} : \left\| \bar{\mathbf{s}}(\mathbf{r})-\boldsymbol{\theta}_u \right\|_{\infty} \leq \epsilon \right\}.
\]
Changing one item response by one point changes its normalized
domain score by \(1/\bigl((L-1)|\mathcal{I}_j|\bigr)\).
Thus, attainable normalized domain scores form a uniform grid over \([0,1]\), and every target score in this interval is within half a grid step of an attainable score.
Since the bounded sampling of Appendix~\ref{app:trait-sampling} guarantees \(\theta_{u,j}\in[l_j,r_j]=[0,1]\), the set \(\mathcal{R}(\boldsymbol{\theta}_u)\) is non-empty whenever \(\epsilon\geq 1/\bigl(2(L-1)|\mathcal{I}_j|\bigr)\) for all \(j\) (\(\epsilon\geq 1/96\) for the BFI-2).
The scoring function depends only on each domain sum.
Therefore, redistributing aligned responses within a domain does not change \(\mathbf{s}(\mathbf{r})\) as long as the sum remains unchanged, so the map \(\mathbf{r}\mapsto\mathbf{s}(\mathbf{r})\) is many-to-one.
The construction is therefore one-to-many by design, and sampling from \(\mathcal{R}(\boldsymbol{\theta}_u)\) selects one of the many behavioral realizations of the same personality trait profile.

\paragraph{Synthesis and verification.}
We synthesize answer sheets by rejection sampling from an LLM-based proposal distribution conditioned on the target scores and the item metadata (item text, facet membership, and keying direction): \[ \mathbf{r}^{(t)} \sim q\!\left( \cdot \mid \boldsymbol{\theta}_u, \{(\mathrm{item}_i, \mathcal{F}(i), k_i)\}_{i=1}^{m} \right), \qquad t=1,2,\ldots \] Each proposal is re-scored with the deterministic scoring function above, and the first proposal satisfying \(\mathbf{r}^{(t)}\in\mathcal{R}(\boldsymbol{\theta}_u)\) is accepted.
Proposals that violate the tolerance are discarded and regenerated, up to a maximum of \(K_{\mathrm{item}}\) attempts.
Because verification uses the same scoring rule as the published instrument, acceptance certifies that the answer sheet's aggregate scores recover \(\boldsymbol{\theta}_u\) within \(\epsilon\).

\subsection{From Answer Sheets to Behavioral Evidence}
The accepted, direction-aligned responses form the item-level response profile \[ \tilde{\mathbf{r}}_u = \left( \tilde{r}_{u,1},\ldots,\tilde{r}_{u,m} \right), \] whose entries specify fine-grained, item-level behavioral tendencies (together with the induced facet scores \(s_{u,j,f}\)).
As described in Sec.~\ref{sec:bp-gen}, these tendencies are translated into the behavior pattern \(\boldsymbol{\pi}_u\), which guides the downstream event generation of Sec.~\ref{sec:memory-gen}.
The translation extracts the salient evidence from \(\tilde{\mathbf{r}}_u\) and compiles it into frequency quotas.

\paragraph{Evidence extraction.}
Each aligned response is recentered around zero to give a signed tendency \[ v_{u,i} = \frac{2\,\tilde{r}_{u,i}-(L+1)}{L-1} \in[-1,1], \] whose sign gives the response direction and whose magnitude gives its salience.
The \emph{evidence set} \(\mathcal{J}_u\) retains at most \(K_{\mathrm{ev}}\) items, namely those of largest salience among the items with \(|v_{u,i}|\geq\tau_{\mathrm{ev}}\).
Each retained item is represented downstream by its item text, its direction, and a strength band \(\mathrm{lvl}(|v_{u,i}|)\in\{1,2,3\}\) assigned by a nondecreasing step map.
All later stages depend on \(\tilde{\mathbf{r}}_u\) only through this evidence.
Therefore different answer sheets in \(\mathcal{R}(\boldsymbol{\theta}_u)\) can retain different evidence, and may induce different behavior patterns from the same personality trait vector.
This extends the one-to-many design above to the behavioral level.

\subsection{Behavior-Pattern Planning and Frequency Guarantees}
We define a fixed set \(\mathcal{C}\) of allowed activity categories.
Each behavior-pattern entry selects one category from \(\mathcal{C}\), and the episodes generated in Appendix~\ref{app:event-tree} use the same category labels.
Conditioned on the evidence, an LLM planner proposes at most \(M\) entries.
Each entry \(g\) specifies an activity category \(a_g\in\mathcal{C}\), a set of allowed days \(\Delta_g\subseteq\{1,\ldots,D\}\), and cited evidence \(\mathcal{J}_g\subseteq\mathcal{J}_u\).
An activity may occur only on days in \(\Delta_g\); a weekly pattern adds every matching day to this set.
An entry has one of two kinds.
A \emph{block} creates standalone timeline episodes.
An \emph{annotation} instead adds an observable natural-language detail \(h_g\) to episodes of an existing block that shares its category and overlaps its days.
This block is called the annotation's \emph{host}; for example, an annotation may add ``initiating small talk'' to a coffee-shop visit.
Proposals that invent categories or omit citations are discarded and regenerated.
When two or more annotations are proposed, they must not all use the same activity category; proposals that violate this diversity rule are also discarded and regenerated.
Annotations without a host block are dropped.
Compilation then assigns each entry its quota deterministically, \[ c_g = \min\Bigl\{ \max\bigl\{1,\ \mathrm{rnd}\bigl(\omega_{\ell(g)}\,|\Delta_g|\bigr)\bigr\},\ \mathrm{cap}(a_g)\,|\Delta_g| \Bigr\}, \] where \(\ell(g)=\max_{i\in\mathcal{J}_g}\mathrm{lvl}(|v_{u,i}|)\) is the strongest cited band, \(\omega_{1}<\omega_{2}<\omega_{3}\) are fixed per-day rates, \(\mathrm{rnd}\) rounds to the nearest integer, and \(\mathrm{cap}:\mathcal{C}\to\mathbb{N}_{>0}\) caps within-day recurrence.
Annotation quotas are further capped by their host's quota.
The behavior pattern is the compiled set \[ \boldsymbol{\pi}_u = \bigl\{ (a_g,\, y_g,\, c_g,\, \Delta_g,\, h_g,\, \operatorname{host}(g)) \bigr\}_{g=1}^{G}, \qquad G\leq M, \] where \(y_g\in\{\textsc{blk},\textsc{ann}\}\) records the entry kind, \(h_g\) is the annotation detail and is empty for blocks, and \(\operatorname{host}(g)\) is the identifier of an annotation's host block and is empty for blocks.
The quota scheduler of Appendix~\ref{app:event-tree} realizes every entry exactly \(c_g\) times on its allowed days, verified from the generated timelines, so the induced daily rates \(\boldsymbol{\nu}_u=(c_1/D,\ldots,c_G/D)\) are recovered exactly by the behavioral-fidelity analysis there.
The profile \(\tilde{\mathbf{r}}_u\) thus serves as the quantitative encoding of \(\boldsymbol{\pi}_u\).

\medskip
\noindent\textbf{Proposition 3} (Effect of response salience on event frequency)\textbf{.}
\emph{Fix an entry \(g\) and hold \(\Delta_g\), \(\mathcal{J}_g\), the
host quota, and all other responses fixed.
If the aligned response \(\tilde{r}_{u,i}\) of a cited item \(i\in\mathcal{J}_g\) moves farther from the neutral response without changing the sign of \(v_{u,i}\), then \(|v_{u,i}|\) increases and the quota \(c_g\), and hence the rate \(\nu_g=c_g/D\), does not decrease.
}

\noindent\emph{Proof sketch.}
The salience \(|v_{u,i}|\) increases in the stated direction, and \(c_g\) is obtained by applying a sequence of nondecreasing operations to it: the band map \(\mathrm{lvl}\), the increasing rates \(\omega\), the clamped rounding above, and the host cap with \(c_{g'}\) held fixed.
\hfill\ensuremath{\square}

\medskip
\noindent
Band discretization and rounding introduce ties while preserving the
guaranteed monotonic relationship.
These ties allow different response configurations within a band to produce the same quota, retaining one-to-many diversity within each band.
The threshold \(\tau_{\mathrm{ev}}\) also defines a monotone eligibility rule: an item cannot support an entry while \(|v_{u,i}|<\tau_{\mathrm{ev}}\), but becomes eligible to support one after crossing the threshold.
Our release sets \(\tau_{\mathrm{ev}}=0.25\) with bands split at \(0.4\) and \(0.7\), \(K_{\mathrm{ev}}=20\), \(M=11\), \(|\mathcal{C}|=426\), \(\mathrm{cap}(\cdot)\in\{1,2,3\}\), and \((\omega_1,\omega_2,\omega_3)=(1/7,2/7,4/7)\) occurrences per allowed day.

\section{Event-Grounded Generation Tree}
\label{app:event-tree}
This section formalizes the \emph{event-grounded generation tree} of Sec.~\ref{sec:memory-gen}, which expands the behavior pattern prior \(\boldsymbol{\pi}_u\) into a long-term episodic history.
Algorithm~\ref{alg:event-tree} summarizes the tree expansion after the quota scheduler has constructed the activity chains.
The key prompts realizing the LLM operators are reproduced in Appendix~\ref{sec:prompts}.

\paragraph{Inputs.}
The generation horizon spans \(D\) consecutive days, organized at the coarse level into calendar months \(\mathcal{M}_1,\ldots,\mathcal{M}_m\) and at the fine level into consecutive windows \(\mathcal{G}_1,\ldots,\mathcal{G}_g\) of \(W\) days each.
For each day \(d\), an activity chain \[ \mathbf{c}_d = \left( e_{d,1},\ldots,e_{d,n_d} \right) \] is constructed from the behavior pattern prior \(\boldsymbol{\pi}_u\) by the quota scheduler below (Fig.~\ref{fig:setoka-pipeline}), where each episode \(e_{d,i}=(a_{d,i},[t^{\mathrm{s}}_{d,i},t^{\mathrm{e}}_{d,i}],\mathcal{H}_{d,i})\) specifies an activity category \(a_{d,i}\in\mathcal{C}\), drawn from the closed catalog of Appendix~\ref{app:scale-mapping}, a time interval, and a possibly empty set \(\mathcal{H}_{d,i}\) of annotation details drawn from the \(h_g\) of the entries hosted by the episode.
A per-user resource pool \(\mathcal{P}\) of contacts, locations, and applications, generated once from the aggregate activity summary (Prompt~\ref{prompt:pool-gen}), constrains all entity mentions so that generated scenes refer to a shared, closed set of entities.
We write \(\Lambda_{\mathrm{anchor}}\), \(\Lambda_{\mathrm{link}}\), \(\Lambda_{\mathrm{bind}}\), and \(\Lambda_{\mathrm{scene}}\) for the LLM operators, each realized by a dedicated prompt.

\paragraph{Quota scheduling.}
The chains are constructed from \(\boldsymbol{\pi}_u\) by a deterministic scheduler rather than sampled freely.
Each day is first initialized with a background template of routine episodes (sleep, meals, work) whose categories lie outside the quota entries.
Block entries are then placed one occurrence at a time.
For each occurrence of entry \(g\), the scheduler first keeps the allowed days in \(\Delta_g\) whose same-category load is below \(\mathrm{cap}(a_g)\).
It chooses the remaining day with the smallest total load, using same-category load and then weekday as tie-breakers.
It then inserts an episode of category \(a_g\) into a free interval on that day.
The capacity bound \(c_g\leq\mathrm{cap}(a_g)|\Delta_g|\) of Appendix~\ref{app:scale-mapping} makes each entry individually placeable, and if several entries jointly exhaust a day's capacity or free intervals, the construction is rejected and rescheduled.
During rescheduling, the scheduler backtracks to the most recent placement with an untried allowed day or free interval and selects the next candidate under the same ordering.
If no candidate remains, it reports that the proposed behavior pattern is infeasible.
Annotation entries are realized afterwards.
For entry \(g\), the scheduler sorts the realized occurrences of \(\operatorname{host}(g)\) by time and selects \(c_g\) approximately equally spaced occurrences.
This spreads the annotations across the horizon.
Each selected episode adds \(h_g\) to its annotation set \(\mathcal{H}_{d,i}\).
A final verification recounts every quota from the finished chains and rejects the construction on any deviation, so every released skeleton realizes each entry exactly \(c_g\) times.

\paragraph{Hierarchical context construction (top-down).}
Higher levels of the tree provide long-range context before any concrete event is written.
For each month \(\mathcal{M}\), an anchor-selection operator picks at most \(\kappa\) memorable episodes, \[ \mathcal{A}_{\mathcal{M}} = \Lambda_{\mathrm{anchor}}\!\left( \{\mathbf{c}_d\}_{d\in\mathcal{M}} \right), \qquad |\mathcal{A}_{\mathcal{M}}|\leq\kappa, \] and the anchor catalog is \(\mathcal{A}=\bigcup_{\mathcal{M}}\mathcal{A}_{\mathcal{M}}\).
Recall bindings tell a later scene to refer to an important episode from another month.
First, the linking operator assigns cross-month relevance weights \(L=\Lambda_{\mathrm{link}}(\mathcal{A})\).
Then, for each half-month period \(p\), the binding operator uses these weights to construct the binding set \[ \mathcal{B}_p = \Lambda_{\mathrm{bind}}(\mathcal{A}, L, p).
\]
Each binding \(b=(e,\alpha,\delta,\chi)\) specifies the target
episode \(e\), the referenced anchor \(\alpha\in\mathcal{A}\), whether
the reference points to the past or future through
\(\delta\in\{\text{past recall},\text{future reference}\}\), and the
content channel \(\chi\) in which the reference should appear, such as a
message or note.
Bindings inject long-range references into subsequent scene expansion, so that events months apart remain mutually consistent without retaining the full history in context.
After the activity chains, anchor catalog, and cross-month weights have been fixed, the operators for different months or half-month periods can run independently in parallel.
We write \(\mathcal{B}=\bigcup_p\mathcal{B}_p\) and \(\mathcal{B}(d)\subseteq\mathcal{B}\) for the bindings whose target episode occurs on day \(d\).

\paragraph{Sequential sibling expansion with bounded memory (bottom level).}
Concrete event scenes are generated window by window.
Disjoint windows are expanded independently in parallel; within a window, days are expanded sequentially under a rolling memory state \(\mathcal{S}\) that summarizes the preceding \(W\) days (per-day briefs, people seen, and locations visited).
For day \(d\), the scene operator receives the day's episodes, the scenes already generated for earlier episodes on the same day \(\mathbf{s}_{\prec}\), the memory state, the day's recall bindings, and the resource pool (Prompt~\ref{prompt:scene-gen}): \[ \mathbf{s}_d = \Lambda_{\mathrm{scene}}\!\left( \mathbf{c}_d, \mathbf{s}_{\prec}, \mathcal{S}, \mathcal{B}(d), \mathcal{P} \right), \] producing exactly one scene per episode, each expressing the annotation details \(\mathcal{H}_{d,i}\) of its episode in the narrative.
Each scene \( s_{d,i} = (\sigma_{d,i}, P_{d,i}, \ell_{d,i}, \mathcal{I}_{d,i}) \) comprises a narrative description \(\sigma\), participants \(P\) drawn from the contact pool, a location \(\ell\) from the location pool, and structured information points \(\mathcal{I}\) that downstream stages use for record derivation and evidence grounding.
Each scene also carries the identity fields of its episode unchanged, namely the activity category, the interval, and the annotation set \(\mathcal{H}_{d,i}\).
After each day, the memory state is updated as \(\mathcal{S}\gets\textsc{Update}(\mathcal{S},\mathbf{s}_d)\), retaining only the most recent \(W\) days, which bounds the context of every LLM call regardless of the horizon length.

\paragraph{Validation and fallback.}
Every scene-level call is validated structurally: the response must contain exactly \(n_d\) scenes, each with a non-empty description and pool-consistent entities.
Invalid responses are retried up to \(K_{\mathrm{scene}}\) times with corrective feedback appended to the prompt.
If a day still fails, its episode list is split into halves: the left half is generated first, and the right half is conditioned on the left half's scenes, recursively, so that sibling continuity is preserved even under fallback.
The recursion stops at single-episode chunks, and a failure at that level raises a generation error.
Our release instantiates \(W=7\), \(\kappa=5\), and \(K_{\mathrm{scene}}=3\).

\paragraph{Record derivation.}
Each scene is finally expanded into the records from which the memory systems construct their memories.
Each of the 23 schemas in Sec.~\ref{sec:experiments} has a derivation rule \(\textsc{Derive}_{\varsigma}\).
The rule copies shared identity fields, including the event identifier, interval, participants \(P\), and location \(\ell\), directly from the scene.
It derives schema-specific fields from the information points \(\mathcal{I}\) and the narrative \(\sigma\).
For example, a purchase information point yields a payment row whose merchant comes from \(\mathcal{I}\), whereas a communication scene yields messages addressed to members of \(P\).
A schema emits records only when the scene contains the information it requires.
Thus, one event usually produces a small complementary set of relational rows, graph edges, and semi-structured documents.
Point-record timestamps remain inside the scene interval, allowing query generation to match each point timestamp to its enclosing interval (Appendix~\ref{sec:query-generation}).
All entity mentions come from the closed resource pool \(\mathcal{P}\).
Because all records for an event derive their shared fields from the same scene and pools, those fields agree by construction.
This gives the single-source-of-truth property of Sec.~\ref{sec:method}.

\paragraph{Output.}
The output consists of the event scene set \(\mathcal{E}_u=\{s_{d,i}\}\) and the heterogeneous records derived from these scenes.
The set covers every episode in the horizon, and each scene carries a unique event identifier.
EM queries expose subsets of each scene's cues as predicates (Appendix~\ref{sec:query-generation}).

\begin{algorithm}[!t]
\caption{Event-Grounded Generation Tree}
\label{alg:event-tree}
\begin{algorithmic}[1]
\Require
Activity chains \(\{\mathbf{c}_d\}_{d=1}^{D}\) scheduled from \(\boldsymbol{\pi}_u\);
months \(\mathcal{M}_1,\ldots,\mathcal{M}_m\);
window size \(W\);
anchor budget \(\kappa\);
retry limit \(K_{\mathrm{scene}}\);
resource pool \(\mathcal{P}\)

\Ensure
Event scene set \(\mathcal{E}_u\); heterogeneous record set \(\mathcal{R}_u\)

\Statex \textit{Phase 1: hierarchical context construction}
\For{each month \(\mathcal{M}\) \textbf{in parallel}}
    \State
    \(\mathcal{A}_{\mathcal{M}}
    \gets
    \Lambda_{\mathrm{anchor}}(\{\mathbf{c}_d\}_{d\in\mathcal{M}})\)
    with
    \(|\mathcal{A}_{\mathcal{M}}|\leq\kappa\)
\EndFor

\State
\(\mathcal{A}
\gets
\bigcup_{\mathcal{M}}\mathcal{A}_{\mathcal{M}}\);
\quad
\(L
\gets
\Lambda_{\mathrm{link}}(\mathcal{A})\)

\For{each half-month period \(p\) \textbf{in parallel}}
    \State
    \(\mathcal{B}_p
    \gets
    \Lambda_{\mathrm{bind}}(\mathcal{A},L,p)\)
\EndFor

\State
\(\mathcal{B}
\gets
\bigcup_p\mathcal{B}_p\)

\Statex \textit{Phase 2: sequential-parallel scene expansion}
\State
\(\mathcal{E}_u\gets\emptyset\)

\For{each window \(\mathcal{G}\) \textbf{in parallel}}
    \State
    \(\mathcal{S}\gets\emptyset\)
    \Comment{rolling memory state}

    \For{each day \(d\in\mathcal{G}\) in temporal order}
        \State
        \(\mathbf{s}_d\gets{}\)\Call{GenScenes}{$\mathbf{c}_d,(\,),\mathcal{S},\mathcal{B}(d),\mathcal{P}$}

        \State
        \(\mathcal{E}_u
        \gets
        \mathcal{E}_u\cup\mathbf{s}_d\)

        \State
        \(\mathcal{S}
        \gets
        \textsc{Update}(\mathcal{S},\mathbf{s}_d)\)
        \Comment{keep last \(W\) days}
    \EndFor
\EndFor

\Statex \textit{Phase 3: record derivation}
\State
\(\mathcal{R}_u\gets\emptyset\)

\For{each scene \(s\in\mathcal{E}_u\) \textbf{in parallel}}
    \For{each schema \(\varsigma\)}
        \State
        \(\mathcal{R}_u
        \gets
        \mathcal{R}_u\cup\textsc{Derive}_{\varsigma}(s,\mathcal{P})\)
        \Comment{zero or more records}
    \EndFor
\EndFor

\State \Return \(\mathcal{E}_u,\;\mathcal{R}_u\)

\Statex
\Function{GenScenes}{$\mathbf{e},\mathbf{s}_{\prec},\mathcal{S},\mathcal{B},\mathcal{P}$}
    \For{\(t=1,\ldots,K_{\mathrm{scene}}\)}
        \State
        \(\mathbf{s}
        \gets
        \Lambda_{\mathrm{scene}}(\mathbf{e},\mathbf{s}_{\prec},\mathcal{S},\mathcal{B},\mathcal{P})\)

        \If{\(|\mathbf{s}|=|\mathbf{e}|\) and every scene in \(\mathbf{s}\) is valid}
            \State \Return \(\mathbf{s}\)
        \EndIf

        \State append corrective feedback to the prompt
    \EndFor

    \If{\(|\mathbf{e}|=1\)}
        \State \textbf{raise} generation failure
    \EndIf

    \State split \(\mathbf{e}\) into halves \(\mathbf{e}_{L},\mathbf{e}_{R}\)

    \State
    \(\mathbf{s}_{L}\gets{}\)\Call{GenScenes}{$\mathbf{e}_{L},\mathbf{s}_{\prec},\mathcal{S},\mathcal{B},\mathcal{P}$}

    \State
    \(\mathbf{s}_{R}\gets{}\)\Call{GenScenes}{$\mathbf{e}_{R},\mathbf{s}_{\prec}\Vert\mathbf{s}_{L},\mathcal{S},\mathcal{B},\mathcal{P}$}

    \State \Return \(\mathbf{s}_{L}\Vert\mathbf{s}_{R}\)
\EndFunction
\end{algorithmic}
\end{algorithm}

\paragraph{Behavioral fidelity.}
We now show that the construction preserves the behavior pattern prior \(\boldsymbol{\pi}_u\) exactly at any horizon, stated through the daily target rates \(\boldsymbol{\nu}_u=(c_1/D,\ldots,c_G/D)\) induced by its quotas (Appendix~\ref{app:scale-mapping}).
Let \(\phi\) map a day's activity chain to the \(G\) behavioral features that \(\boldsymbol{\pi}_u\) constrains.
The \(g\)-th component counts the episodes of that day realizing entry \(g\), namely episodes of category \(a_g\) placed for a block entry, and episodes whose annotation set contains \(h_g\) for an annotation entry.
The skeleton statistic is \(\Phi^{\mathrm{skel}}_D(\{\mathbf{c}_d\})=\tfrac{1}{D}\sum_{d}\phi(\mathbf{c}_d)\), and the history statistic is \[ \Phi_D(H) = \frac{1}{D}\sum_{d=1}^{D}\phi\bigl(\hat{\mathbf{c}}_d(H)\bigr), \] where \(\hat{\mathbf{c}}_d(H)\) is the day-\(d\) chain recovered from the expanded history \(H\) by collecting each scene's identity fields, namely its activity category, its interval, and its annotation set.

\medskip
\noindent\textbf{Proposition 4} (Preservation of behavioral statistics during expansion)\textbf{.}
\emph{Conditional on successful validation, the reconstruction is the identity, \(\hat{\mathbf{c}}_d(H)=\mathbf{c}_d\) for every day \(d\), and hence \(\Phi_D(H)=\Phi^{\mathrm{skel}}_D(\{\mathbf{c}_d\})\).}

\noindent\emph{Proof sketch.}
The scene operator emits exactly one scene for each episode, and the validation-and-fallback step enforces this count or raises a generation error.
Each scene copies the episode's activity category, interval, and annotation set unchanged.
Therefore, reading these fields from the scenes reconstructs the original chain exactly:
\[
\hat{\mathbf{c}}_d(H)=\mathbf{c}_d.
\tag*{\(\square\)}
\]

\medskip
\noindent\textbf{Theorem 1} (Exact behavioral fidelity)\textbf{.}
\emph{Every history \(H\) that passes the quota-scheduler verification and the expansion validation satisfies, for every horizon length \(D\),}
\[
\Phi_D(H)\;=\;\boldsymbol{\nu}_u.
\]

\noindent\emph{Proof sketch.}
The scheduler verification recounts every quota from the finished chains and rejects any deviation, so each entry \(g\) is realized exactly \(c_g\) times across the skeleton and \(\Phi^{\mathrm{skel}}_D(\{\mathbf{c}_d\})=(c_1/D,\ldots,c_G/D)=\boldsymbol{\nu}_u\) by the definition of \(\phi\).
Proposition~4 transfers the identity to the expanded history.
\hfill\ensuremath{\square}

\medskip
\noindent
Two consequences are worth noting.
First, fidelity does not degrade with the horizon.
The quota construction guarantees that the behavioral statistic equals \(\boldsymbol{\nu}_u\) for every \(D\).
In contrast, under autoregressive generation with a bounded context window, the deviation from \(\boldsymbol{\nu}_u\) can grow with the horizon after early constraints leave the window.
For comparison, suppose a relaxed scheduler generated days independently, with expected daily feature vector \(\boldsymbol{\nu}_u\) and bounded feature counts.
A standard concentration bound would then give \(\lVert\Phi_D(H)-\boldsymbol{\nu}_u\rVert_\infty= O\bigl(\sqrt{\log(G/\delta)/D}\bigr)\) with probability at least \(1-\delta\), where \(\delta\) is the failure probability.
The exact-quota construction therefore gives a strictly stronger guarantee.
Second, Proposition~4 shows that scene expansion preserves each day's behavioral fields exactly.
Therefore, behavioral fidelity does not require context older than \(W\) days.

\section{Query Generation}
\label{sec:query-generation}
This appendix describes in detail how we generate the queries introduced in Sec.~\ref{sec:query-gen}.
We use three generation routes.
SM and multi-hop factual queries are generated by enumerating templates over the record schema and instantiating them with stratified query sampling.
EM queries use a separate partial-cue generator.
For BP and PT, we construct questions whose reference answers come from ground truth recorded at generation time.
Finally, an LLM converts every query into natural language.

\begin{figure*}[t]
\centering
\includegraphics[width=0.71\linewidth]{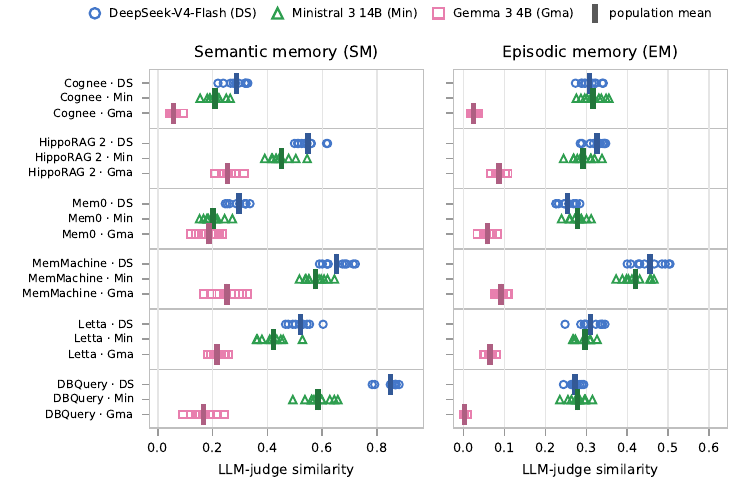}
\renewcommand{\thefigure}{F.1}%
\caption{\textbf{Per-user LLM-judge similarity on the SM and EM levels}.
Each hollow marker shows the mean for one user under one system-backbone combination.
The thick tick shows the population mean reported in Table~\ref{tab:main-results}.
Per-user scores cluster around the mean of each system-backbone combination, whereas the means differ substantially across combinations.
}
\label{fig:exp-user-var}
\end{figure*}

\textbf{Template enumeration.}
We represent the 23 schemas (Table~\ref{tab:schemas}) as a graph (Fig.~\ref{fig:schema-graph}).
Each node is a table or collection, and each edge represents one of two join types.
A foreign-key equi-join connects records within a data model.
A temporal point-in-interval join connects a point event, such as a payment, photo, or health sample, to the session whose interval contains the event timestamp; this join can connect different data models.
We use depth-first search to enumerate simple paths containing \(k\in\{1,2,3,4\}\) tables.

Each template selects one field on the path as the answer and \(k'\in\{1,2,3\}\) other fields as predicates.
Key fields and fields used only to connect tables are excluded.
Each table is labeled as structured, text, or graph, corresponding to the relational, semi-structured, and graph data models of Sec.~\ref{sec:method}.
We call the multiset of labels along a path its \emph{modality signature}; repeated labels are counted.
The value \(k\) is the join depth reported in Appendix~\ref{app:dataset-stats}.
Thus, \(k=1\) yields single-record SM templates, whereas \(k>1\) yields the MH templates analyzed in Fig.~\ref{fig:exp-heterogeneity}.

\textbf{Instantiation and validation.}
Each template is instantiated by binding its predicate fields to values drawn from the corpus and then computing the resulting answer.
The instance must pass three checks.
First, \emph{replay match} requires that re-executing the binding reproduce the same answer, which rules out predicates that are ambiguous under the sampled data.
Second, \emph{minimality} requires that no predicate can be dropped without changing the answer.
Third, \emph{true multimodality} requires that, for a template spanning multiple modalities, every modality on the path affect the answer.
Removing any one of them must either change the answer or make the target unreachable.
This ensures a template is never credited with a modality it does not actually depend on.

\textbf{Stratified query sampling.}
Naively sampled queries could make join depth or modality signature correlated with unrelated properties, such as the number of valid answers or the age of the supporting evidence.
We therefore divide candidates into buckets defined by modality signature, join depth, and query-field count and assign equal generation quotas to these buckets.
Within each bucket, we select queries so that the distribution of the unrelated properties matches the corpus-wide average.
This control helps prevent those properties from confounding comparisons across join depths.

\textbf{Episodic queries.}
An EM query tests whether a model can recover a missing part of an event from partial information, following the setting of Sec.~\ref{sec:query-gen}.
The generator selects an event scene, provides some of its time, participant, and location cues as predicates, and withholds one cue as the answer.

\textbf{Semanticization.}
For each validated instance, we first create a structured, SQL-like representation of the retrieval task.
This representation specifies which records must be connected, which conditions select the relevant records, and which field should be returned as the answer.
An LLM rewrites this representation into a single natural-language question that preserves the exact retrieval target and filtering condition without naming SQL constructs (Prompt~\ref{prompt:semanticization}); a second LLM pass validates that the rewritten question still targets the same fields, and items that fail are discarded.
This keeps queries natural for the evaluated memory-augmented agents while retaining, for every item, a machine-checkable template against which evidence and answers are automatically graded.
Applied to our release, this procedure yields 868 factual-QA templates, instantiated into 9{,}898 factual-QA items over the 10 users; the dedicated episodic generator contributes a further 4{,}203 episodic-reconstruction items (Table~\ref{tab:ds-queries}).

\textbf{Behavior pattern and personality trait queries.}
Template enumeration is designed for factual queries whose answer is a field reached along a schema path.
BP and PT queries instead test aggregation and abstraction, respectively, so we construct them directly from the ground truth recorded during data generation.
A BP query aggregates a user's episodic records within an activity category and asks for a behavioral statistic, such as how often the activity occurs.
The statistic computed from the generated event history is the reference answer (Sec.~\ref{sec:query-gen}).
A PT query asks the model to estimate one personality trait for one user.
For each trait, we rank the 10 users by their predicted scores and compare this ranking with the ranking induced by the ground-truth vectors \(\boldsymbol{\theta}_u\), using Kendall's \(\tau\) as defined in Sec.~\ref{sec:metrics}.
The released BP and PT sets contain 110 and 50 items (Table~\ref{tab:ds-queries}), namely one per behavior pattern and one per trait dimension for each of the 10 users.

\section{Additional Experimental Results}
\label{sec:appendix}
\setcounter{figure}{1}%
This appendix provides supporting results referenced from Sec.~\ref{sec:experiments}: cross-user variability and reproducibility of the main results (Figs.~\ref{fig:exp-user-var} and~\ref{fig:exp-repro}), heterogeneity as a difficulty axis (Figs.~\ref{fig:exp-heterogeneity} and~\ref{fig:exp-depth-ansrate}), and a per-trait analysis of the personality-trait level (Fig.~\ref{fig:exp-pt-tau}).

\subsection{Cross-User Variability and Reproducibility}

\label{app:robustness}

\begin{figure}[!t]
\centering
\includegraphics[width=0.98\linewidth]{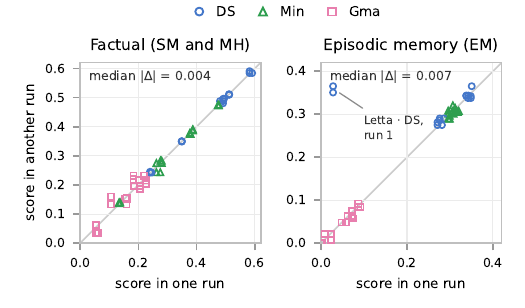}
\caption{Run-to-run reproducibility across 18 system-backbone configurations, with three runs per configuration.
Each point compares the cell-level scores of two runs, over all run pairs.
}
\label{fig:exp-repro}
\end{figure}

\begin{figure}[!t]
\centering
\includegraphics[width=0.98\linewidth]{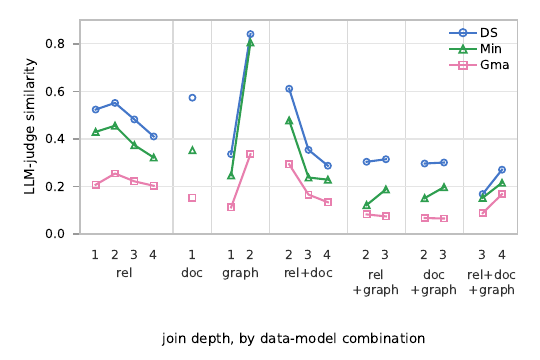}
\caption{LLM-judge similarity of the factual queries by the data-model combination of their evidence and join depth, averaged over the six systems and 10 users.
Combinations follow Fig.~\ref{fig:queries-by-depth}, so queries whose join chain spans several data models are marked explicitly.}
\label{fig:exp-heterogeneity}
\end{figure}

\begin{figure*}[!t]
\centering
\includegraphics[width=0.8\linewidth]{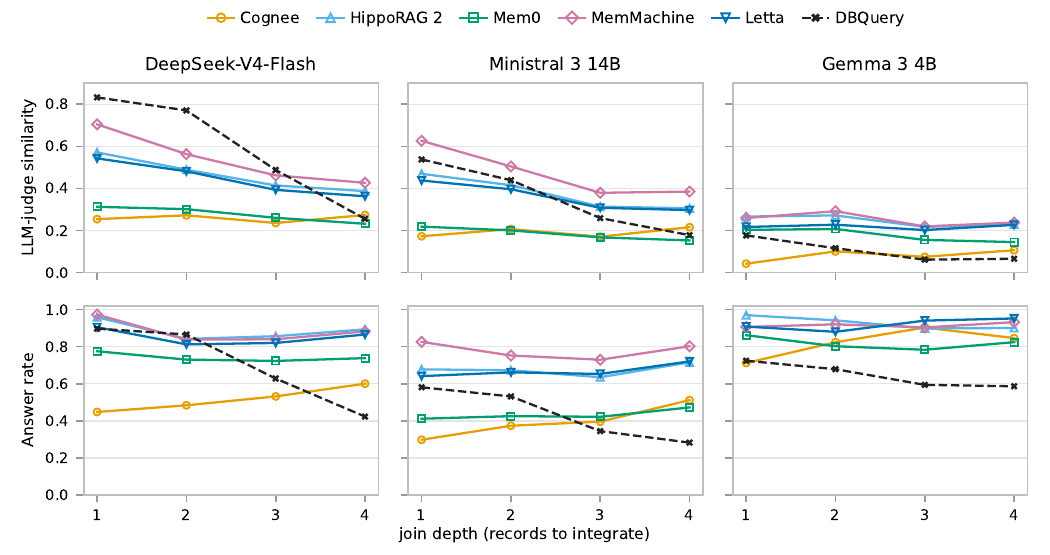}
\caption{LLM-judge similarity (top row) and answer rate (bottom row) of the factual queries by join depth, shown per system and backbone and pooled over the three data models and the 10 users.
The dashed line marks the memory-free DBQuery baseline.}
\label{fig:exp-depth-ansrate}
\end{figure*}

\begin{figure}[!t]
\centering
\includegraphics[width=0.88\linewidth]{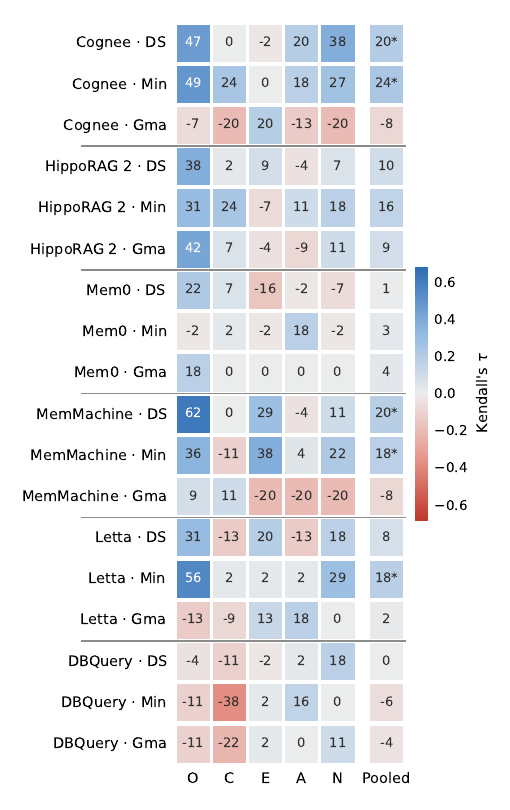}
\caption{Kendall's $\tau$ ($\times 100$) between the predicted and ground-truth orderings of the 10 users, per Big-Five trait dimension (O, C, E, A, N) and pooled over the five dimensions (rightmost column, as in Table~\ref{tab:main-results}).
Blue indicates agreement with the ground-truth ordering, red an inverted ordering, gray chance level; a star marks pooled scores significant at $p<.05$ under a permutation test.
}
\label{fig:exp-pt-tau}
\end{figure}

Figure~\ref{fig:exp-user-var} disaggregates the SM and EM columns of Table~\ref{tab:main-results} by user.
For every system-backbone combination, the cross-user standard deviation remains below \(0.05\) on the \([0,1]\) judge-score scale.
This variation is about one order of magnitude smaller than the differences between systems or backbones, indicating that individual synthetic users do not dominate the main comparisons.
We also reran all system-backbone configurations twice.
For each pair of runs we take the absolute difference of the two cell-level scores, and report the median over the 18 configurations and the three run pairs.
This median shift is \(0.004\) on the factual queries and \(0.007\) on EM.
Both shifts are far smaller than the gaps between systems and backbones in Table~\ref{tab:main-results}, indicating that run-to-run sampling noise does not affect the reported comparisons.
Figure~\ref{fig:exp-repro} visualizes this comparison by plotting the cell-level scores of the three runs against each other.
Nearly all points lie on the diagonal.
The one exception is Letta$\times$DeepSeek on EM, where run 1 scores \(0.03\) while runs 2 and 3 score \(0.35\) and \(0.37\).
The two later runs agree with each other, so this deviation reflects an occasional run-level failure of the memory system rather than noise in the judged answers.

\subsection{Heterogeneity as a Difficulty Axis}
\label{app:heterogeneity}

For each backbone, Figure~\ref{fig:exp-heterogeneity} decomposes the accuracy of the factual (SM and MH) queries by the data-model combination of their evidence and by join depth (Appendix~\ref{app:dataset-stats}).
We report the LLM-judge similarity averaged over the six systems (the five memory systems and the DBQuery baseline) and the 10 users.
Two patterns emerge.
First, within the combinations that span several depths, accuracy declines as more records must be integrated, although not always monotonically.
For DeepSeek, purely relational accuracy falls from \(0.52\) at depth 1 to \(0.41\) at depth 4, and relational-plus-document accuracy from \(0.61\) at depth 2 to \(0.29\) at depth 4.
Second, at a fixed depth the combinations that involve the contact graph are the hardest.
Under DeepSeek, every graph-involving combination stays below \(0.32\) at depths 2-4, whereas the pure graph queries of depth 2, which look up the relationship between two contacts, reach \(0.84\).
Attribute lookups on a single contact node score only \(0.34\), which suggests that identifier-valued attributes such as email addresses and phone numbers are often lost during memory construction.
Thus, even among factual queries, integrating more records and crossing data models both introduce additional difficulty.

Figure~\ref{fig:exp-depth-ansrate} decomposes the same factual pool by system rather than by data model, and pairs the accuracy at each join depth with the corresponding answer rate.
Comparing accuracy with answer rate helps explain why performance decreases as join depth increases.
A system may continue answering at the same rate but make more mistakes, or it may abstain from more queries.
The five memory systems follow the first pattern.
Their answer rates stay roughly flat across depths under all three backbones, and for Cognee they even rise, so their accuracy decline in the top row reflects wrong answers rather than growing abstention.
DBQuery instead follows the second pattern.
Its answer rate falls from \(0.90\) at depth 1 to \(0.42\) at depth 4 with DeepSeek, and from \(0.58\) to \(0.28\) with Ministral, so the SQL baseline responds to deeper joins by abstaining.
This contrast extends the calibration finding of Sec.~\ref{sec:experiments} within a single query level.
Abstention tracks difficulty only for the memory-free baseline, whereas memory systems answer deep-join queries as readily as direct lookups.

\subsection{Personality-Trait-Level Analysis}
\label{app:pt-analysis}

Figure~\ref{fig:exp-pt-tau} decomposes the PT column of Table~\ref{tab:main-results} by trait dimension.
Under a permutation test on the pooled $\tau$, only five of the 18 configurations rank the users significantly better than chance: Cognee$\times$Ministral ($\tau{=}0.24$, $p{=}.002$), Cognee$\times$DeepSeek ($\tau{=}0.20$, $p{=}.006$), MemMachine$\times$DeepSeek ($\tau{=}0.20$, $p{=}.028$), Letta$\times$Ministral ($\tau{=}0.18$, $p{=}.029$), and MemMachine$\times$Ministral ($\tau{=}0.18$, $p{=}.034$).
None of the remaining configurations, including every Gemma configuration, performs significantly better than chance under this test.
Most of the observed rank-order agreement comes from openness, whose per-trait \(\tau\) reaches \(0.62\).
A possible explanation is that activities associated with openness, such as reading, media use, and learning, leave comparatively dense evidence in the corpus.
The PT analysis yields two conclusions.
First, even the configurations that perform significantly better than chance recover only weak user rankings, with \(\tau\leq 0.24\).
Second, per-answer judge similarity can overestimate PT performance.
A predictor that assigns the same mid-scale value to every user may receive substantial per-answer similarity, but it cannot distinguish among users and therefore has Kendall's \(\tau=0\).
We therefore report Kendall's \(\tau\) for PT in Table~\ref{tab:main-results}.

\section{Prompts}
\label{sec:prompts}
We present the key prompts used by the generation pipeline (Sec.~\ref{sec:method}) and the evaluation harness (Sec.~\ref{sec:metrics}).
Long field enumerations are abbreviated with ``[\ldots]'', and bracketed notes summarize content that is assembled programmatically (pools, schemas, retrieved evidence).

\begin{promptbox}[label={prompt:pool-gen}]{Stage-4 Pool Generation}
You are a behavioral data scientist generating realistic personal data pools for a simulated smartphone user.

Your task is to create apps, contacts, and locations that can support later event generation from the supplied timeline summary.

All generated pool items must be consistent with the label summary.
A pool lists items that later stages may use; it does not assign an item to a particular event. Include enough plausible items to support different events rather than generating only the minimum required.

Language policy:
- Use English for app names, contact display_name, relationship values, location_name, and semantic labels.
- Contact display_name must be English (stable role labels for close family, e.g., Wife, Father; personal names otherwise). [...]
- relationship must be a single English noun: family, friend, colleague, classmate, or service.

[User message provides: home city, per-label summaries (event_count, duration_minutes, covered_day_count, annotation_hints), and the target JSON pool schema for apps/contacts/locations.]
\end{promptbox}

\begin{promptbox}[label={prompt:scene-gen}]{Stage-4 Event Scene Generation}
Generate Event Scene objects for one day of timeline episodes.

IMPORTANT:
- Return exactly N event_scene object(s), one per episode below, in the same order.
- Do not merge episodes; do not split one episode into multiple objects; do not skip vague episodes.
- Follow the output JSON schema; descriptions in English; build structure first, then write description.

Build Protocol (per episode):
1. Treat the source label and annotation as mandatory constraints on the event.
2. Choose scenes[0] for the main action domain.
3. Add scenes[1:] only for distinct same-time actions.
4. Choose people_present and location.
5. Fill compact information_points.
6. Write description from the completed structure.

Source Constraints: the current label and annotation have the highest priority. You may use the full-day context, memory, pools, and recall information to add details to a broad label, but the resulting scene must still describe the event specified by that label and annotation.

Map each activity phrase to an event type as follows. Food or drink maps to meal. Cleaning, repair, laundry, or pet care maps to household. Washing, grooming, or medication maps to personal_care. A phone call, message, or other call maps to communication. ``Travel related to ...'' maps to travel. Reading, watching, or writing must use the most suitable type among work_study, media_leisure, game, and digital_file. Creative output maps to creative_hobby. [...]

People and Location: people_present holds exact Contacts Pool display_name values for direct participants only. Location prefers exact Location Pool names; travel uses {"location_type":"move","from_name":...,"to_name":...}. [...]

Information Points contain compact facts that identify the event and may later be recovered from database evidence, such as its subtype, main object or topic, merchant, app or platform, recipient, and whether a corresponding record should be created. Named apps must use exact Apps Pool names. Do not repeat the time, location, or people. [...]

[Prompt also includes: full-day timeline overview, previously generated scenes, episodes to generate, the contacts/location/apps pools, short-term scene memory, recall bindings, and the output JSON schema.]
\end{promptbox}

\begin{promptbox}[label={prompt:semanticization}]{QA Semanticization}
Rewrite SQL-like QA items as concise natural-language retrieval requests. Do not answer the query; describe what information the SQL is intended to retrieve.

Rules:
- Preserve the exact retrieval target and filtering meaning.
- Use plain, natural English; do not mention SQL keywords (SELECT, FROM, WHERE, JOIN, DISTINCT).
- Do not invent extra conditions, entities, or answer values.
- Keep each description to one sentence; prefer task phrasing (Find ..., Retrieve ..., Return ...).

Return ONLY valid JSON: {"items":[{"qa_id":"...","sql_semantic":"..."}]}
\end{promptbox}

\begin{promptbox}[label={prompt:answerer}]{Answerer (Memory-Augmented QA Agent)}
You are an extractive database QA assistant. Extract the answer only from the retrieved evidence provided by the evaluation harness. The harness retrieves memory using the original question and then supplies that evidence to you.
- The answer must come verbatim from the retrieved evidence: do not rewrite, translate, explain, infer, or reformat.
- The user message provides answer_format and answer_json_shape; when answer is not null, it must match that JSON structure exactly, with no extra fields.
- String answers: copy the original string verbatim (case, punctuation, spacing). List answers: copy each element verbatim. Numeric answers: preserve the numeric value.
- If no retrieved evidence directly supports an answer, answer must be null and confidence 0.
- Output only valid JSON with fields: answer, confidence, evidence_notes.
\end{promptbox}

\begin{promptbox}[label={prompt:judge}]{LLM-as-a-Judge}
[System] You are a strict QA evaluator. Score semantic equivalence between a gold answer and a model answer. Return only a JSON object with keys similarity and reason; similarity must be a number from 0 to 1.
- Use 1.0 only when the answers are fully equivalent, including entities, numbers, dates, and units.
- Use 0.7-0.9 for mostly correct answers with minor formatting or harmless extra detail.
- Use 0.3-0.6 for partially correct answers missing important details.
- Use 0.0-0.2 for wrong, contradictory, empty, or unsupported answers.

[User] {"question": ..., "gold_answer": ..., "model_answer": ..., "instructions": "Judge only whether model_answer answers the question equivalently to gold_answer."}
\end{promptbox}

\FloatBarrier
\bibliographystyleapp{aaai2027}
\bibliographyapp{reference/survey,reference/psychology,reference/memory-benchmark,reference/llm-table-understanding,reference/related-work}

\end{document}